\pdfoutput=1
\documentclass[11pt]{article}

\newif{\ifhidecomments}

\ifhidecomments
    \newcommand{\hossein}[1]{}
\else
    \newcommand{\hossein}[1]{\textcolor{purple}{[#1 ---\textsc{hossein}]}}
\fi

\ifhidecomments
    \newcommand{\bashar}[1]{}
\else
    \newcommand{\bashar}[1]{\textcolor{blue}{[#1 ---\textsc{bashar}]}}
\fi

\ifhidecomments
    \newcommand{\ke}[1]{}
\else
    \newcommand{\ke}[1]{\textcolor{red}{[#1 ---\textsc{ke}]}}
\fi

\ifhidecomments
    \newcommand{\shihao}[1]{}
\else
    \newcommand{\shihao}[1]{\textcolor{pink}{[#1 ---\textsc{shihao}]}}
\fi

\ifhidecomments
    \newcommand{\joel}[1]{}
\else
    \newcommand{\joel}[1]{\textcolor{orange}{[#1 ---\textsc{joel}]}}
\fi

\ifhidecomments
    \newcommand{\alex}[1]{}
\else
    \newcommand{\alex}[1]{\textcolor{green}{[#1 ---\textsc{alex}]}}
\fi

\usepackage[]{EMNLP2022}

\usepackage{times}
\usepackage{latexsym}

\usepackage[T1]{fontenc}
\usepackage[utf8]{inputenc}

\usepackage{graphicx}
\usepackage{enumitem}
\setitemize{noitemsep,topsep=0pt,parsep=0pt,partopsep=0pt}
\newcommand{\datasetname}{CrisisLTLSum}
\usepackage{microtype}

\usepackage{inconsolata}
\usepackage{algorithm}
\usepackage{algorithmic}
\usepackage{hyperref}

\title{\datasetname: A Benchmark for Local Crisis Event Timeline Extraction and Summarization}
\usepackage{multirow}
\usepackage{amsmath}
\usepackage{bbding}
\usepackage{pifont}
\usepackage{amssymb}%

\newcommand{\cmark}{\ding{51}}%
\newcommand{\xmark}{\ding{55}}%

\newcommand{\train}{{\sc Train}}
\newcommand{\dev}{{\sc Dev}}
\newcommand{\test}{{\sc Test}}

\author{Hossein Rajaby Faghihi$^{1}${\Thanks{{\hspace{2pt}Work done as Research Interns at Dataminr, Inc.}}} , Bashar Alhafni$^{2*}$,
Ke Zhang$^{3}$, Shihao Ran$^{3}$, \\\textbf{Joel Tetreault$^{3}$}, \textbf{Alejandro Jaimes$^{3}$}\\
  $^{1}$Michigan State University, $^{2}$New York University Abu Dhabi, \\$^{3}$Dataminr, Inc.\\
  \texttt{rajabyfa@msu.edu,alhafni@nyu.edu},\\\texttt{\{kzhang,sran,jtetreault,ajaimes\}@dataminr.com}
  }

\begin{document}
\maketitle
\begin{abstract}
Social media has increasingly played a key role in emergency response: first responders can use public posts to better react to ongoing crisis events and deploy the necessary resources where they are most needed. Timeline extraction and abstractive summarization are critical technical tasks to leverage large numbers of social media posts about events. Unfortunately, there are few datasets for benchmarking technical approaches for those tasks. This paper presents \datasetname{}, the largest dataset of local crisis event timelines available to date. \datasetname{} contains 1,000 crisis event timelines across four domains: wildfires, local fires, traffic, and storms. We built \datasetname{} using a semi-automated cluster-then-refine approach to collect data from the public Twitter stream. 
Our initial experiments indicate a significant gap between the performance of strong baselines compared to the human performance on both tasks.
Our dataset, code, and models are publicly available.\footnote{ \url{https://github.com/CrisisLTLSum/CrisisTimelines}}
% \joel{do you mean to say that there is a significant gap between human performance and system performance, showing this to be a challenging task?}\hossein{Yes} 
\end{abstract}

\section{Introduction}
\begin{figure}[t]
\centering
\includegraphics[width=\linewidth]{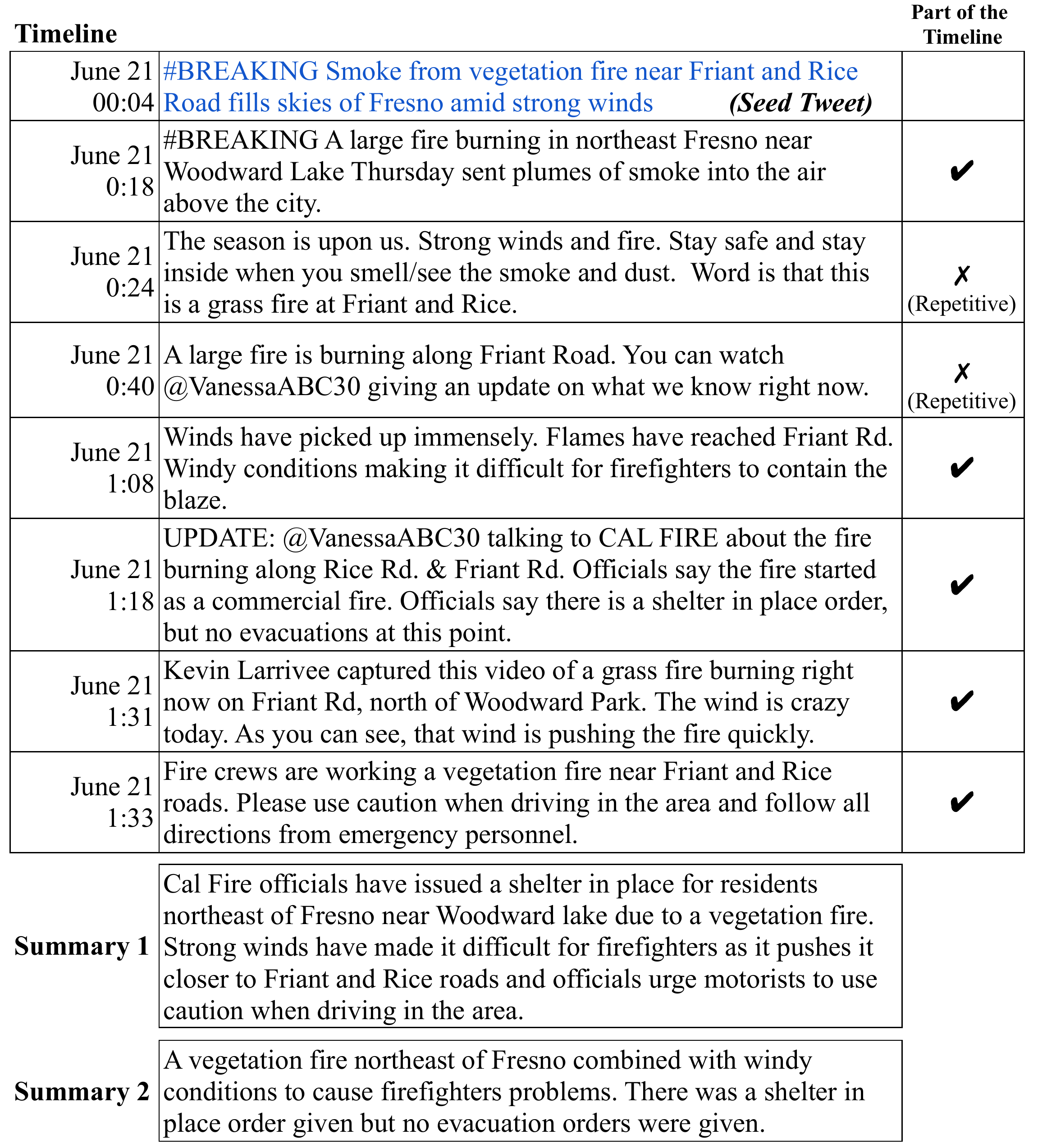}
\caption{This is a sample annotated timeline from \datasetname{}. The noisy timeline is the set of tweets sorted chronologically. The first tweet is the seed of the event. \cmark~means that the tweet is annotated to be part of the timeline and \xmark~indicates that the tweet is excluded. The reason for the exclusion is written under the mark.}
\label{fig:main-example}
\end{figure}

We present \datasetname{}, the first dataset on extraction and summarization of local crisis event timelines from Twitter. An example of an annotated timeline in \datasetname{} is shown in Figure \ref{fig:main-example}. A timeline is a chronologically sorted set of posts, where each brings in new information or updates about an ongoing event~(such as a fire, storm, or traffic incident). \datasetname{} supports two complex downstream tasks: \textbf{timeline extraction} and \textbf{timeline summarization}. 
As shown in Figure \ref{fig:main-example}, the {\bf timeline extraction} task is formalized as: given a seed tweet as the initial mention of a crisis event, extract relevant tweets with updates on the same event from the incoming noisy tweet stream. This task is crucial for real-time event tracking. 
The {\bf timeline summarization} task aims to
generate abstractive summaries of evolving events by aggregating important details from temporal and incremental information. 
% The summary tends to provide a quick and up-to-date overview to stakeholders (e.g., first responders) on what has been happening during the crisis event.\hossein{The last sentence can be removed}

\datasetname{} can facilitate research in two directions: 1) NLP for Social Good (crisis domain), and 2) natural language inference and generation, i.e., timeline extraction and summarization tasks.
Here, we discuss the importance and the differences of \datasetname{} compared to previous work for both of these aspects.
% This is for the importance of humanitarian aid and our difference
Towards the first direction, the extraction of real-time crisis-relevant information from microblogs~\cite{zhang2019tracking,mamo2021fine} plays a vital role in providing time-sensitive information to help first responders understand ongoing situations and plan relief efforts accordingly~\cite{sarter1991situation}. 
\datasetname{} goes beyond the task of categorizing each single crisis-relevant post independently~\cite{imran2013practical,olteanu2014crisislex,imran2016twitter,alam2018crisismmd,wiegmann2020disaster,alam2021humaid,alam2021crisisbench} and enables a more challenging task for extracting new updates of an ongoing crisis event from incoming posts and summarizing them with respect to the important event details. This can help provide time-sensitive updates while avoiding missing critical information in the bulk of the posts in microblogs due to the high volume of redundant and noisy information~\cite{alam2021humaid}. To the best of our knowledge, this is the first annotated dataset for such an extraction task, while this problem has been tackled before in unsupervised settings~\cite{zhang2018geoburst}.
% Compared to prior research in this domain which has mostly focused on extracting crisis-relevant tweets from significant large-scale disaster events and mapping them to different categorizations such as humanitarian types~\cite{wiegmann2020disaster,imran2013practical,olteanu2014crisislex,imran2016twitter,alam2018crisismmd,alam2021crisisbench,alam2021humaid}, we go beyond this task by enabling a novel and more challenging task for extracting new updates of an ongoing crisis event from incoming posts
% Recent research has heavily investigated this direction in both real-time~\cite{schulz2013see,rudra2018extracting,kumar2011tweettracker} and retrospective manner~\cite{repp2018extracting,alam2021crisisbench,alam2021humaid}. Existing datasets mainly focus on extracting all crisis-relevant tweets about significant large-scale disaster events and mapping them to humanitarian categories~\cite{wiegmann2020disaster,imran2013practical,olteanu2014crisislex,imran2016twitter,alam2018crisismmd,alam2021crisisbench,alam2021humaid}, actionable information~\cite{mccreadie2019trec}, or specific witness levels~\cite{zahra2020automatic}.
% \datasetname{} goes beyond the task of categorizing each single post independently and enables a novel and more challenging task of extracting new updates of an ongoing crisis event from incoming posts. 

% is hard even for human due to the distributed nature of information in social media

% This is for local focus
Moreover, we focus on the extraction of \textbf{local} crisis events. The term ``local'' indicates that an event is bound to an exact location, such as a building, a street, or a county, and usually lasts for a short period. 
% This is different from prior work that focused on large-scale and long-lasting crisis events affecting multiple and broader geographical areas (such as Hurricane Sandy). 
Building a corpus of local crisis events is particularly useful for first responders but also challenging because the timelines of these events are often not captured in existing knowledge sources. This means one has to design mechanisms for automatically detecting and tracking events directly from the Twitter stream, which is especially hard for existing clustering methods~\cite{guille2015event,asgari2021topic} given the low number of available tweets for each local event.

For the second point, \datasetname{} enables NLP research on the complex tasks of timeline extraction and abstractive summarization. 
These tasks are particularly challenging in the context of social media. First, the process of identifying and extracting relevant updates for a specific event has to contend with the large volume of noise~\cite{alam2021humaid} and informal tone~\cite{rudra2018extracting} compared to other domains such as news. Additionally, summarizing an on-going event helps toward a quick and better understanding of its progress. This requires a good level of abstraction with important details covered and properly presented (e.g., the temporal order of event evolution). 
%for several reasons. 
% The first challenge is the need to identify relevant updates from noisy tweets stream, and aggregate information from multiple tweets~(compared to extractive summarization~\cite{yan2011evolutionary,tran2015timeline}) while considering temporal information. 
% Second, compared to other sources of information~\cite{barros2019natsum,chen2019-ijcai}, summarizing timelines extracted from Twitter is more complicated due to the amount of noise~\cite{alam2021humaid}, the informal language~\cite{rudra2018extracting}, and the distributed nature of information compared to centralized sources such as news~\cite{nallapati2016abstractive}.
\datasetname{} is the first dataset to provide human-written timeline summaries to support research in this direction. 
% on the crisis domain or extracted information from Twitter in general. 

% This task is essentially a more challenging multi-document summarization(MDS)~\cite{} as it requires further integration of temporal consideration in the summarization process. Lastly, \datasetname{} focuses on abstractive summarization. Except for a few~\cite{}, most of the prior research~\cite{} on timeline summarization has taken an extractive approach to form summaries by gathering representative sentences from important data points. The abstractive summarization task is more challenging due to the need for restructuring and generating new sentences to capture the important information. Compared to the existing abstractive summarization on other domains~\cite{}, summarizing Twitter timelines is more challenging due to the amount of noise~\cite{alam2021humaid}, the informal language~\cite{rudra2018extracting}, and the distributed nature of information compared to centralized sources such as news~\cite{nallapati2016abstractive}. \datasetname{} is the first dataset to provide human-written gold timeline summaries on the crisis domain or on top of Twitter in general.
\datasetname{} is developed through a two-step semi-automated process to create 1,000 local crisis timelines from the public Twitter stream. To our best knowledge, this is the first timeline dataset focusing on ``local'' crisis events with the largest number of unique events.   %%\cite{10.1007/978-3-030-72113-8_33}. Joel - removing this 
The contributions of this paper are as follows:
\begin{itemize}
    \item We propose \datasetname{}, which is the largest dataset over
    % clean and concrete 
    local crisis event timelines.
    % \joel{what does clean and concrete mean?}\hossein{We have specified what is redundant, repetitive, and made them into concrete timeline updates as new tweets come} 
    Notably, this is the first benchmark for abstractive timeline summarization in the crisis domain or on Twitter.
    \item We develop strong baselines for both tasks, and our experiments show a considerable gap between these models and human performance, indicating the importance of this dataset for enabling future research on extracting timelines of
    % clean and concrete
    crisis event updates and summarizing them.
\end{itemize}

% In the remainder of this paper, we discuss the automated timeline generation in Section \ref{sec:generate}, the post-processing and annotation platform on generated timelines \ref{sec:correct}, and the baselines and experiments in Section \ref{sec:experiments}. 

\section{Related Work}
Our work in this paper is related to two main directions of crisis domain datasets for NLP and timeline summarization.

\textbf{Crisis Datasets for NLP:} Prior research has investigated generating datasets from online social media (e.g., Twitter) on large scale crisis events, while providing labels for event categories~\cite{wiegmann2020disaster,imran2013practical}, humanitarian types and sub-types~\cite{olteanu2014crisislex,imran2016twitter,alam2018crisismmd,wiegmann2020disaster,arachie2020unsupervised,alam2021humaid,alam2021crisisbench}, actionable information~\cite{mccreadie2019trec}, or witness levels~\cite{zahra2020automatic} of each crisis related post. 
% flow and finding new information about ongoing events. 
% While these datasets enable simple classification tasks, our \datasetname{} 
While existing datasets on crisis event timelines \cite{binh2013predicting, tran2015timeline, 10.1007/978-3-030-72113-8_33} are limited to a small set of large-scale events, 
\datasetname{} covers a thousand timelines compared to only tens of events covered by each of the existing datasets. Additionally, we further go beyond the simple tweet categorization by enabling the extraction of information that include updates over the events' progress.
% \joel{the extraction of updating information sounds a little off;  what are you trying to say exactly?}\hossein{I think it was updated information, someone changed it to updating. We mean extracting new information that provide an update about the previous events.}

\textbf{Timeline Summarization:} 
Timeline summarization (TLS) was firstly proposed in~\citet{allan2001temporal}, which extracts a single sentence from the news stream of an event topic. In general, the TLS task aims to summarize the target's evolution (e.g., a topic or an entity) in a timeline~\cite{martschat2018temporally,ghalandari2020examining}. 
% The task has been investigated in two directions of extractive and abstractive approaches. 
% talk about most of the existing work focuses on extractive method 
Existing approaches of TLS are mainly based on \textbf{extractive} methods, which are often grouped into several categories. For instance, Update Summarization~\cite{dang2008overview,li2009enhancing} aims to update the previous summary given new information at a later time, while Timeline Generation~\cite{yan2011evolutionary,tran2015timeline,martschat2018temporally} aims to generate itemized summaries as the timeline, where each item is extracted by finding important temporal points (e.g., spikes, changes or clusters) and selecting representative sentences. Another category, Temporal Summarization, was first proposed in the TREC shared task~\cite{aslam2013trec} with follow-up work~\cite{kedzie2015predicting}, which targets extracting sentences from a large volume of news streams and social media posts as updates for large events. Temporal Summarization is close to the first task (Timeline Extraction) proposed in \datasetname{}.

There have been a few recent works on \textbf{abstractive} timeline summarization across different domains, e.g., biography \cite{chen2019-ijcai}, narratives \cite{barros2019natsum}, and news headlines \cite{steen-markert-2019-abstractive}, where the human-written summaries are directly collected from the web.  The abstractive summarization goal is to generate a set of sentences summarizing the context of interest without taking the exact words or phrases from the original text but rather by combining them and summarizing the important content.
To our best knowledge, \datasetname{} is the first to provide human-written summaries for crisis event timelines collected from noisy social media stream. Recent research~\cite{nguyen2018tsix} has also investigated the summarization task based on tweets in other domains which essentially do not reflect the challenges in the summarization of an evolving event.

\section{\datasetname{} Collection}
\label{sec:collection}
This section presents our semi-automated approach to collect \datasetname{}. We first extract clusters of tweets as noisy timelines and then refine them via human annotation to get clean timelines that only include non-redundant, informative, and relevant tweets.
% This approach relies on an automated noisy timeline extraction from the Twitter stream and human effort to further clean the timelines to only include non-redundant, informative, and relevant information. \ke{this is already described later, let us remove}  
% First, we apply an automated noisy timeline collection pipeline to collect clusters of tweets related to each local event. Second, we manually select clusters as noisy timelines which contain interesting crisis event stories.
% Third, we recruit high-quality crowd-workers to refine selected clusters by extracting relevant, non-redundant, and informative tweets to generate clean timelines. Annotators further write summaries to describe how the event progresses over time.
% \datasetname{} includes 1000 local crisis event timelines covering four domains of wildfire, local fire, traffic, and storm, collected across seven states in the USA and two states of Australia during 2020 and 2021. 

\subsection{Noisy Timeline Collection}
\label{sec:generate}
\begin{figure}[t]
    \centering
    \includegraphics[width=\linewidth]{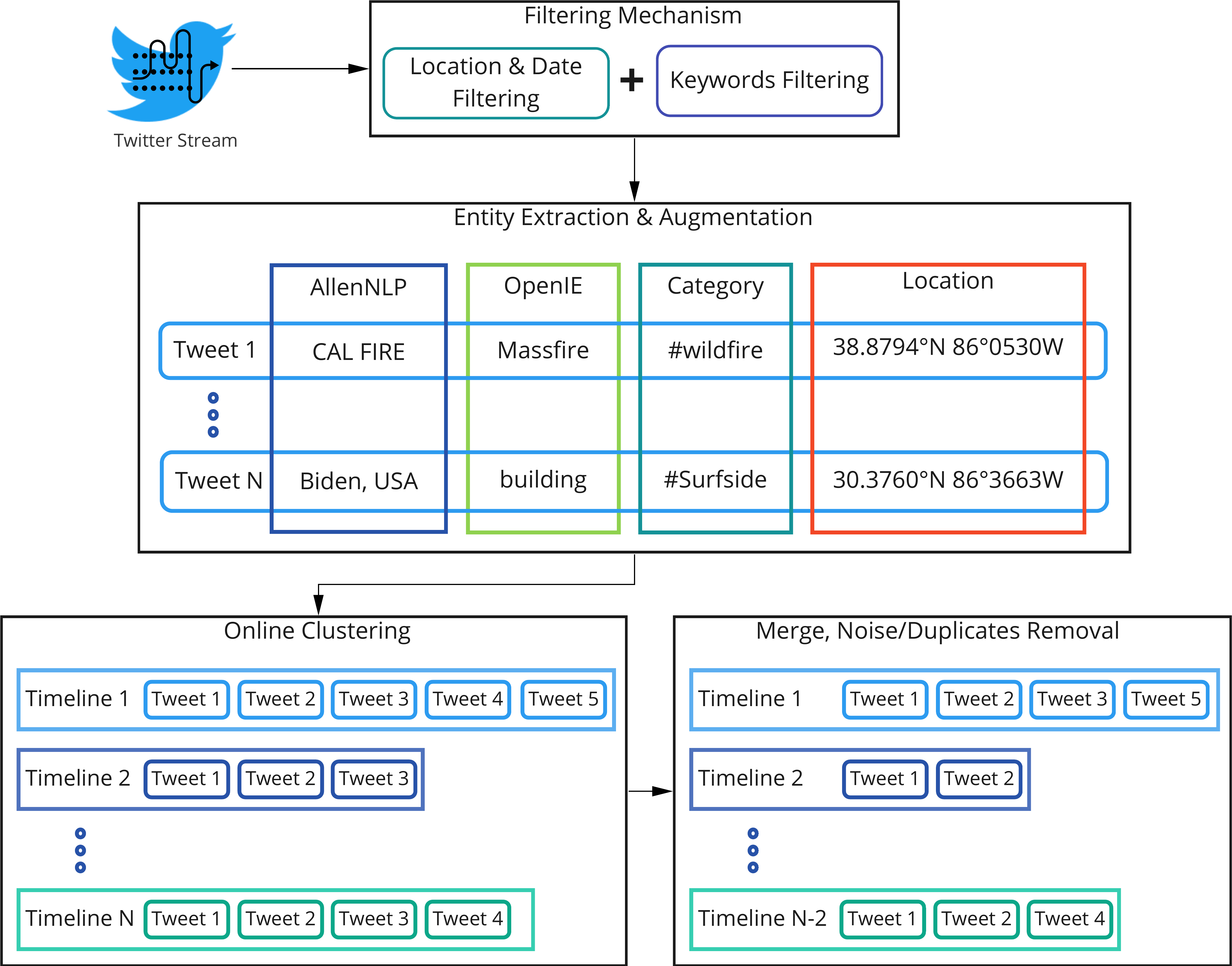}
    \caption{
    % This figure shows the pipeline of the online clustering, starting from the Twitter stream. 
    The process of noisy timeline collection. 
    % includes initial filtering methods, entity extraction, location augmentation, online clustering, merging, and noise \& duplicate removal. 
    The output of this step are noisy clusters which are used to create the dataset.}
    \label{fig:model}
\end{figure}

%%% Overview of the process
Figure \ref{fig:model} shows the process for generating a set of noisy timelines starting from the Twitter stream and followed by pre-processing and knowledge enhancement steps, the online clustering method, and post-processing \& cleaning steps. 
% \iffalse

% \fi

% \ke{This paragraph can be replaced with a figure to who the flow of data collection. or just removed} The process starts with filtering the tweets' stream to find tweets that are both relevant to a location and a category of interest. Since we limit these two properties of tweets in the initial step, we have to repeat our generation pipeline for each disaster category(e.g., fire, storm) and location of interest(e.g., California, New York). Next, we use a set of modules to extract and augment important entities of each tweet. Finally, we perform online clustering to find groups of tweets related to the same local event. The clustering method utilizes a custom similarity metric which we will discuss more in Section \ref{sec:online}. Additionally, we execute merging, noise removal, and de-duplication methods to improve the quality of the generated clusters. 
% The goal of these post-processing steps are to ensure that the clusters are not solely the repetition of the same information and contain informative tweets about the crisis events. We borrow and customize the definition of informative tweets from the existing research on disaster events on twitter~\cite{alam2021humaid}. Here, an informative tweet should contain a solid report of a crisis event's progress or share a non-personal fact about the ongoing actions. We also do not consider prayers, charity work, or fundraising as informative information since we are more interested in studying the flow of the actual crisis event.

\paragraph{Location, Time, and Keywords Filtering}
We limit the incoming tweets to specific geographical areas, periods, and domains of interest. 

The location filtering relies on a list of location candidates created by gathering cities, towns, and famous neighborhoods in a big area of interest. A tweet is considered relevant to our area of interest if 1) the text mentions one of the candidates, 2) the geo-tag matches the area of interest, or 3) the user location matches the area of interest.
To limit the tweets to a specified crisis domain, we curate domain-specific keywords and only select tweets with phrases matching one of the keywords.
This approach is not comprehensive or exhaustive but somewhat representative of each crisis domain. Improving this method to be more encompassing is an area for future research.
The combinations of (area $a$, domain $d$, time $t$) are manually selected so that the events of type $d$ at location $a$ are more frequent during time period $t$. For instance, wildfire events are most likely to happen in California from May through August, while the same type of event is more likely from December to Match in Victoria~(Australia). More details with examples of curated keywords can be found in Appendix \ref{sec:app_collection}.

% This filtering step identifies a possible set of tweets about a location of interest during a specified period.
% For each location (usually a state), first, we create a set of location candidates from the possible location mentions, including cities, towns, and famous neighborhoods. Next, we check whether any of these candidate locations appear in the tweets' text or the location tag of the tweet object. We further include all the tweets from users whose user-location tag is in the locations candidate list.
% To limit the date, we run the online clustering methods for each month of data separately. As a result, each initial filtered tweet set can be flagged by Location\_$l$\_Year\_$y$\_Month\_$m$.

\paragraph{Entity Extraction}
% different location extraction methods
This step aims to extract entity mentions from the tweet text and provide additional information that can be used to help identify related tweets.
We use three different modules. First, we use a pre-trained neural model
% \footnote{\url{https://storage.googleapis.com/allennlp-public-models/ner-elmo.2021-02-12.tar.gz}} 
from AllenNLP~\cite{gardner_allennlp}, trained on CoNLL03~\cite{tjong_2003_conll}, to extract entities with types of people, location, and organization. 
Although this module extracts some important entities in the text, it fails to extract uncommon entities or special mentions such as the name of a wildfire. To address this, similar to prior research~\cite{zheng2020srlgrn}, we further exploit the extractions from OpenIE~\cite{Stanovsky2018SupervisedOI} and select the noun arguments with less than ten characters as entities. Lastly, we add the tweet's hashtags to the entity set. 
Since location mentions are crucial in extracting local events and existing models have low performance detecting them from noisy tweets text, we further developed a BERT-based NER model tuned on Twitter data to detect location mentions. 

% The location mentions are critical in generating local timeline clusters. However, the accuracy of pre-trained models is pretty low for extracting location mention from tweets' text. Here, we create a rule-based location detector fitted to the Twitter data and further train a custom model to extract location entities on a small annotated set. Although both of these models~(especially the rule-based) may add false-positive mentions to the locations list, later, we prune these mentions using the location augmentation module.

\paragraph{Location Augmentation}
% mapping locations to physical objects with coordinates
% The goal of location augmentation is to map the extracted location mentions to physical locations.
% First, we concatenate the location mentions by the same order appearing
% that appear as a connected sequence 
% in the text and consider them as a single location. 
% Next, while limiting the search to a specific area~(i.e., a US State) of interest,
We use OpenStreetMap API to map location mentions to physical addresses.\footnote{\url{https://www.openstreetmap.org/}} This step provides complementary information about each location while reducing the noise introduced by the entity extraction module through removing location mentions that are wrongly detected or are not located in the area of interest. This is especially important since our focus is on local events happening at specific locations.
% Since OpenStreetMap API matches an input string with a list of possible physical locations with probabilities; we employ a fuzzy procedure to remove or keep the locations based on those probabilities. 

\paragraph{Online Clustering}
\label{sec:online}
% how to cluster tweets as they come into the system
This step aims to mimic the real-life scenario where tweets are sequentially fed into a clustering algorithm~\cite{6871372}. We further choose this method since this is a lot faster than the retro-respective~(all data available at the same time) clustering methods for a large pool of input data. Here, the clustering objective is to group tweets related to the same local event, such as a ``fire in building A'' or a ``wildfire in a specific area''. The online clustering method utilizes a custom similarity metric that combines the similarity of the entities, the closeness of locations in the real world, and the existence of shared hashtags. 
\begin{algorithm}[h]
\caption{Find Similarity of tweet $t_{i}$ and $t_{j}$}
\begin{algorithmic} 
\STATE $similarity = 0$
\IF{$t_1.hashtags \cap t_2.hashtags \neq \emptyset$}
\STATE $similarity = similarity + s_{hashtag}$
\ENDIF
\STATE $dist_{min} = smallest\_distance(t_1, t_2)$
\IF{$dist_{min} \ge max_{dist}$}
\RETURN 0
\ELSIF{$dist_{min} \le min_{dist}$}
\STATE $similarity = similarity + s_{dist}$
\ENDIF
\STATE $top\_pairs =$  find\_matching\_entities$(t_1, t_2)$
% \STATE $norm\_factor = \\ \min(len(t_1$.entities$), len(t_2$.entities$), top\_pairs)$
\STATE $norm\_factor = $
\STATE $\min(len(t_1$.entities$), len(t_2$.entities$), top\_pairs)$
\STATE $s_{entity} = \sum top\_pairs.similarity / norm\_factor$
\STATE $similarity = similarity + s_{entity}$
\RETURN $similarity$
\end{algorithmic}
\label{algo:similarity}
\end{algorithm}
Algorithm \ref{algo:similarity} shows the similarity computation between two tweets.
The \textit{smallest\_distance} computes the minimum physical distance between location mentions given their augmented real-world location~(the output of the location augmentation step). As the distance between higher-level location mentions such as state/city/country is always zero, we simply ignore those location types. The  \textit{find\_matching\_entities} function follows the ideas in \citet{faghihi2020latent} on creating a unique matching matrix, which we use for extracting the top matching pairs of entities from tweets. Here, each entity can only be paired once with the highest matching-score entity from the other tweet. The $min_{dist}$, $max_{dist}$, $s_{hashtag}$, and $s_{dist}$ are hyper-parameters of the clustering algorithm. We have only used heuristics and a small set of executions to tune these hyper-parameters.

The pre-processed set of tweets is passed to the online clustering method, one tweet at a time. For each new tweet, similarity scores are computed between the new tweet and all cluster heads. The new tweet is added to the highest matching-score cluster where the similarity score is higher than $sim_{threshold}$ and the time elapsed between the new tweet and the last update of the cluster is less than $time_{threshold}$. If the previous criteria are met for none of the clusters, a new cluster is created based on the new tweet. During this process, we remove inactive clusters whose last update was at least $expiration_{threshold}$ minutes ago and have less than $tweet_{threshold}$ number of tweets available. A cluster head is always the tweet with the most entity mentions; In case of a tie, the more recent tweet becomes the head of the cluster. The hyper-parameters of this method is noted in Appendix \ref{sec:app_collection}. 

\subparagraph{Cluster Post-Processing}
We apply three post-processing steps to improve the quality of the generated clusters. First, we manually merge pairs of clusters with a cluster head similarity higher than a threshold $head_{min}$. This step compensates for some of the errors from missing entities in the pre-processing step, which affects the intermediate similarity scores in the clustering algorithm.
Second, we use a simple fuzzy sequence-matching technique to remove identical or similar tweets inside each cluster. Third, we train a BERT-based~\cite{devlin-etal-2019-bert} binary classifier to detect informative content, which can be used to prune out the noisy tweets that do not include crisis-relevant information. This classifier is trained on the available labeled data~\cite{alam2021humaid} on tweets' informativeness. Since most of the available tweets in \citet{alam2021humaid} are specific to storm and wildfire domains and there are no representative subsets for our other domains of interest~(traffic, local fire), we only apply this step to the clusters that are generated for those categories. These post-processing steps aim not to prune out all the noisy information but rather to provide a better starting point for our next steps.

\subsection{\datasetname{} Human Annotation}
\label{sec:annotation}
Taking the noisy timelines generated from the previous step, we leverage human annotations to refine and generate clean timelines and summarize them. We, authors of this work, manually selected ~1,000 clusters that contain enough tweets describing how a crisis event evolves,
while specifying the ``seed tweet'' (i.e. the first observed post that describes the ongoing event) of each timeline. 
The detailed process is presented in Appendix \ref{sec:cluster_selection}.
% In particular, we first identify a ``seed tweet'' as the first observed post mentioning a local crisis event, then check whether the following tweets in the cluster contain updates about the same event. Taking the example in Figure \ref{fig:main-example}, we first see ``a vegetation fire happened'' in the seed tweet, then see ``plumes of smoke'', ``windy condition impacts fire control''.
The selected clusters cover events mainly from four crisis domains, including wildfire, local fire, storm, and traffic. More data statistics are shared in Section \ref{sec:analysis}. 

\paragraph{Procedure} We use the Amazon Mechanical
Turk (MTurk) platform to label and refine the noisy clusters to generate a clean timeline and collect the summaries. We split the annotation into multiple batches of Human Intelligence Tasks (HITs), where each batch contains timelines from the same domain.
% \hossein{This is just the same numbers as the other paper, this is not true for us} \ke{updated the text. It seems we have a variant batch size}. 
Each HIT contains three noisy timelines, and we collect annotations from 3 different workers on each. The workers are given the seed tweet and the subsequent tweets sorted by time, and they were asked to read the tweet one by one and answer i) whether the tweet should be part of the timeline, and ii) what is the reason if not. 

A tweet is labeled as part of the timeline only if it satisfies all the following three conditions:
\begin{itemize}
    \item {\em relevant}: talks about the same event indicated in the seed tweet 
    \item {\em informative}: provides facts about the event but not only contains personal points of view
    \item {\em not repetitive}: brings in new information about the ongoing event
\end{itemize}
After reviewing all the tweets, the worker is finally asked to write a concise summary to describe how the event progresses over time. 
Detailed instructions and annotation workflows are presented as Figures \ref{fig:TS_task_page1}-\ref{fig:TS_task_page4_all}
% \ref{fig:TS_task_page2_upper},
% \ref{fig:TS_task_page2_middle}, \ref{fig:TS_task_page2_lower},
% \ref{fig:TS_task_page3_upper},
% \ref{fig:TS_task_page3_lower},  
% \ref{fig:TS_task_page4_all} 
in Appendix \ref{app:sec:anno}.

\paragraph{Annotation Workflow \& Quality Control}
Following prior quality control practices \cite{briakou2021ola}, we use multiple quality control~(QC) steps to ensure the recruitment of high-quality annotators. First, we use location restriction~(QC1) to limit the pool of workers to countries where native English speakers are most likely to be found. Next, we recruit annotators who pass our qualification test~(QC2), where we ask them to annotate 3 timelines. We run several small pilot tasks, each with a replication factor of nine. We check annotators' performance on timeline extraction task against experts' labels and have experts manually review~(QC3) annotators' summary quality. 
Only workers passing all the quality control steps contribute to the final task. 
% multiple  Third, we use \textbf{manual summary qualification~(QC3)} to go over the summaries generated on the annotators on the qualification task and remove those with low-quality summaries.
During the final task, we perform regular quality checks~(QC4), and only use workers who consistently perform well.
%put workers in blacklist if they don't continuously perform well. 
% to repeat the same process as Q3 on on the final annotation task. 

% \ke{this is post-eval, will be presented in quality analysis}
% We further use \textbf{expert check~(QC5)} to evaluate the performance of some workers against 10\% of timelines annotated by experts.
% Finally, we conduct a post-processing \textbf{summary quality evaluation~(QC6)} to evaluate the summary qualities against a set of baseline models in four different dimensions~\cite{viv-2022}. We discuss the results of these experiments and quality evaluations in Section \ref{}.

\paragraph{Compensation}
We compensate the workers at a rate of \$3 per HIT for the task. Each batch of tasks is followed by a one-time bonus that makes the final rate over \$10 per hour.

\section{\datasetname{} Statistics \& Analysis}
\label{sec:analysis}

In this section, we cover comprehensive statistics and analysis of \datasetname{} to further elaborate on the statistical characteristics of our dataset.

\begin{table}[t!]
\centering
\setlength{\tabcolsep}{5pt}
\begin{tabular}{|l|l|l||l|l|}
\hline
\bf{Domain} & \bf Timelines & \bf Tweets & \ding{51} & \ding{55} \\\hline
\bf{Wildfire} & 423 & 4,829  & 1,961 & 2,868\\
\bf{Traffic} & 287  & 2,340 & 831  & 1,509\\
\bf{Fire} & 155  & 1,469  & 640 & 829\\
\bf{Storm} & 109 & 1,767 & 789 & 978  \\
\bf{Other} & 26 & 205  & 82 & 123\\\hline
 \multicolumn{1}{c|}{}& 1,000 & 10,610 & 4,303 & 6,307   \\\cline{2-5}
                        
\end{tabular}

\caption{Data statistics across different crisis domains in terms of the number of timelines and tweets. \ding{51} indicates tweets that are part of the timeline and \ding{55} indicates tweets that are not part of the timeline.}
\label{tab:stats}
\end{table}

\subsection{Dataset Statistics} \label{sec:datastats}
% From the annotated timelines, 45 (4.3\%) are dropped due to annotation errors, leaving 1,000 timelines (10,610 tweets) in \datasetname{}.
Out of these 1,000 annotated timelines~(10,610 tweets) in \datasetname{}, 423~(42\%) are about wildfire, 287~(29\%) are about traffic, 155~(15\%) are about local fire, 109~(11\%) are about storm, and 26~(3\%) are about other types of events~(e.g., building collapse). 
% In total, there are 10,672 tweets across all 1,000 timelines. 
To make the ground-truth on whether each tweet is part of the timeline or not, we take the majority label among the three workers. 4,303~(41\%) tweets are labeled as part of the timeline, whereas 6,307~(59\%) are not. Out of all timelines, 110~(11\%) only include tweets that are part of the timeline, whereas 68~(7\%) did not include any. Table~\ref{tab:stats} presents the statistics across different event crisis domains.

\subsection{Dataset Analysis}
\label{sec:data-analysis}
\begin{figure}[t]
\centering
\includegraphics[width=\linewidth]{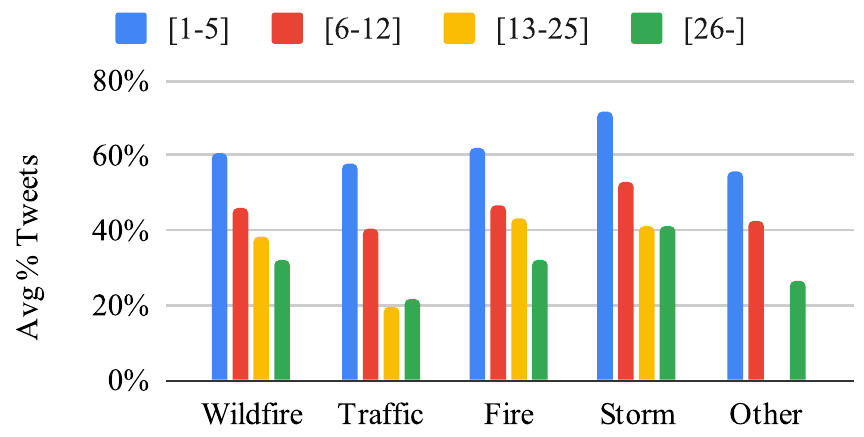}
\caption{Average \% of tweets that are part of the timelines based on the aggregated timelines lengths across different crisis domains.}
\label{fig:stat-fig}
\end{figure}

\paragraph{Timelines by Lengths} Table~\ref{tab:stats-lengths} presents the aggregated length (i.e., number of tweets) distribution of the timelines. The majority of the timelines (447 or 45\%) have five tweets or less. This is observed across all crisis domains except for storm events, where the majority of the timelines (43 out of 109) are 6 to 12 tweets in length. It is worth noting that our dataset includes long timelines of 26 or more tweets, constituting 9\% (94 timelines) across all domains. Figure~\ref{fig:stat-fig} presents the average percentage of part-of-timeline tweets based on the aggregated length distributions of the timelines. We notice an interesting trend across the domains: the longer the timeline, the lower the average percentage of tweets to be part of the timeline.
% This is true for all domains except for traffic, where the timelines with 26 and more tweets have a higher percentage of relevant 

% Timelines length distribution
% What kind of analysis should we do about summaries?

\paragraph{Annotation Quality}
% 1) Best two workers agreement on 1000 timelines -- done
% 2) Timeline Extraction QE on 100 timelines -- done (expert vs majority vote)
% 3) Timeline Summarization QE on 100 timelines
To measure the agreement rate between workers on the timeline extraction task, we consider the annotations of two out of three workers who agree most per timeline. The average timeline-level agreement between those two workers is \textbf{90.06\%}. We also explore a more profound analysis by comparing the workers' annotations on 20\% of the timelines against the annotations of experts. To do so, we evaluate the majority label provided by the three workers against the label provided by the experts. The average timeline-level agreement rate of this analysis is \textbf{91.77\%}.

% Table \ref{} shows the quality of the annotations based on two metrics. The first metric simply reflects the percentage of consistency between three workers of each timeline on the timeline extraction task. Another deeper analysis of the annotation quality is computed by evaluating the workers' annotation against the 10\% expert annotated timelines. Here, the majority vote of the workers, which is the final label of this dataset, is evaluated against the label suggested by experts. 

\subsection{Dataset Splits}

To aid reproducibility when using our dataset for various research experiments, we divided our dataset into: training ({\train}: 70\% or 706 timelines), development ({\dev} 10\% or 86 timelines), and testing ({\test} 20\% or 208 timelines) splits. The splits are created via stratified sampling based on the event crisis domains and the length of timelines~(i.e., number of tweets) as shown in Table~\ref{tab:stats-lengths}. 
% For example, the wildfire timelines containing 1 to 5 tweets, would be randomly split into 70\% for {\train}, 10\% for {\dev}, and 20\% for {\test}. 
Detailed statistics are available in Appendix~\ref{sec:data-splits}.

% We organized the data based on the different crisis event domains and the length of timelines (i.e., number of tweets) per domain.

\begin{table}[t!]
\centering
\setlength{\tabcolsep}{4pt}

\begin{tabular}{|l|l|l|l|l|}
\cline{1-5}
\multirow{2}{*}{\bf Domain} & \multicolumn{4}{|c|}{\bf Aggregated Timelines Lengths}\\\cline{2-5}
& [1 -- 5] & [6--12] & [13--25] & [26 --] \\\hline
\bf{Wildfire} & 175 & 140  & 59 & 49\\
\bf{Traffic} & 158  & 95 & 20  & 14\\
\bf{Fire} & 76  & 45  & 22 & 12\\
\bf{Storm} & 21 & 43 & 27 & 18  \\
\bf{Other} & 17 & 8  & 0 & 1\\\hline
\multicolumn{1}{c|}{} & 447 & 331 & 128 & 94\\\cline{2-5}
                        
\end{tabular}

\caption{Dataset statistics based on the aggregated timelines lengths (i.e., number of tweets) across different crisis domains.}
\label{tab:stats-lengths}
\end{table}

\section{Experiments}
\label{sec:experiments}
Here, we first formalize the definitions of the downstream tasks. We then propose a set of existing naive and advanced models to serve as a baseline for these tasks. Next, we analyze the performance of the baselines on each task and provide error analysis for the best performing baseline to indicate the remaining major challenges of this dataset for future research. 
% Experiments and the splits

\subsection{Task Definitions}
Given a crisis event timeline consisting of $n$ tweets $T = [t_0, t_1, ..., t_n]$, where $t_0$ is the seed tweet, i.e. the first observed post that describes the ongoing event, we define two tasks: \textbf{Timeline Extraction} and \textbf{Timeline Summarization}.

\textbf{Timeline Extraction: }  
% about the event described in the timeline. 
Given an initial seed tweet $t_0$, and following chronologically sorted tweets $t_i, i=1, ... ,n$, the goal of the timeline extraction task is to determine whether tweet $t_i$ is part of the timeline, i.e., related to the same event and adding new information compared to all prior tweets $[t_0, ..., t_{i-1}]$. Up to time $n$, a clean timeline $T_{extracted} = [t_{0}, t_{e}^1, t_{e}^2, .., t_{e}^m],  m \leq n$ is extracted to describe how the event progresses over time.

% timeline $T_{initial} = [t_0, t_1, ..., t_n]$, the goal of the timeline extraction task is to find the timely ordered subset of tweets in $T_{initial}$ where each tweet is related to the same event in previous tweets and contains new information. More formally, the task aims to construct $T_{extracted} = [t_{0}, t_{e}^1, t_{e}^2, .., t_{e}^m]$ from $T_{initial}$ by detecting whether each tweet $t_{i}$ in $T_{initial}$ is relevant, informative, and non-repetitive, hence part of the timeline $T_{extracted}$. In addition to detecting whether each $t_i$ is part of the timeline, a complimentary task is to predict why it is not part of the timeline. The reasons for it to not be part of the timeline are: 1) not related; 2) not informative; and 3) no new information.

%  $T_{initial} = [t_1, t_2, ..., t_n]$, where $t_1$ is the seed of the timeline with $n$ tweets. The goal is to find the timely ordered subset where each next tweet is related to the previous tweets and brings a new information. We denote this set by $T_{answer} = [t_{1}, t_{a}^1, t_{a}^2, .., t_{a}^m]$. 
% To get to $T_{answer}$ from $T_{initial}$, the task is to detect whether each tweet $t_{i}$ belongs to the timeline $[t_{1},..., t_{a}^j]$ where $t_a^j$ appears before $t_i$ in the $T_{initial}$. Additional to detecting whether each $t_i$ belongs to the timeline, a complimentary task is to predict why it does not belong to the timeline. The reasons for not belonging to the timeline are 1) not being related, 2) not being informative, and 3) not having new information.

\textbf{Timeline Summarization: } Similar to previous work by~\newcite{chen2019-ijcai}, this task aims to generate a summary $\hat{Y}=\{\hat{w}_1, ..., \hat{w}_k\}$ with sequence of words $\hat{w}_i$ to 
concisely describe the crisis event and its evolution given the output from the timeline extraction task $T_{extracted}$. In particular, it tries to optimize the parameters that maximize the probability $P(Y/T_{extracted})$ given ground-truth summaries 
$Y=\{w_1, ..., w_k\}$.   
% Our definition of this task is similar to the work by \newcite{chen2019-ijcai}.
% \joel{do we describe their definition, or ours?  If someone is unfamiliar with Chen2019 they will feel they are missing out.  Best to just say what the definition is }
% However, the main difference is that they do not refine the timeline before the summarization process. 

\subsection{Modeling}

\subsubsection{Timeline Extraction}
\paragraph{Naive Baseline} We employ a simple and naive baseline that assigns the label of the majority class observed in {\train} to detect whether a tweet is part of the timeline or not.

\paragraph{Sequence Classification} We leverage pretrained language models to build sentence-level classifiers. 
% Given a timeline $T = [t_{0}, t_{1}, .., t_{n}]$, where $t_{0}$ is the seed tweet, the goal is to detect whether tweet $t_{i}$ is part of the timeline or not. 
We construct a list of tweet sequences $S$ by concatenating all the tweets that are part of the timeline from $t_{1}$ to $t_i$: $S = [t_{1}, t_{1}+t_{2}, ..., t_{1}+...+t_{n}]$. Therefore, each timeline of $n$ tweets would yield $n$ sequences. For our first sentence-level classification model, we fine-tune BERT~\cite{devlin-etal-2019-bert} on the tweet sequences where every training example would have the form of ``$t_0$ \texttt{[SEP]} $s_i$''; where $s_i \in S$. 
% Given that some of the timelines in our dataset consist of a large number of tweets, the BERT-based classification model has a limitation when processing long tweet sequences due to its maximum sequence length of 512 wordpiece tokens. To address this limitation, we build a second sentence-level classification model by fine-tuning Longformer
% Due to the limitation of pre-trained BERT-based models on encoding long sequences, we propose a second sentence-level classification model by fine-tuning Longformer~\cite{longformer-beltagy}. 
We fine-tune this model using Hugging Face's Transformers~\cite{wolf-etal-2020-transformers}.

% \ke{inference should be clear}
% During inference, we follow the same procedure as in fine-tuning. That is, when predicting whether tweet $t_i$ is part of the timeline, we construct the tweet sequence by concatenating $t_i$ to all the tweets that are before $t_{i}$ and are predicted to be part of the timeline.

\paragraph{Sequence Labeling} Given the sequential nature of timelines, we treat the timeline extraction task as a sequence labeling problem. We create a model by adding a GRU~\cite{cho-etal-2014-learning} on top of BERT. Given a timeline  $T = [t_{0}, t_{1}, t_{2}, .., t_{n}]$, we feed every tweet $t_{i}$ to BERT and get its contextual representation corresponding to the \texttt{[CLS]} token. We then concatenate the representations of all the tweets and feed them to the GRU to predict if each tweet is part of the timeline or not.\footnote{It is important to note that the loss corresponding to the seed tweet $t_0$ is masked out during training.}

\subsubsection{Timeline Summarization}
\paragraph{Naive Baselines} We define three simple models as the first set of timeline summarization baselines: 1) \textit{first-tweet}: uses the first tweet of the timeline as the output summary; 2) \textit{last-tweet}: uses the last tweet of the timeline as the output summary; and 3) \textit{random-tweet}: uses a random tweet from the timeline as the output summary.

\paragraph{Seq2Seq Models} We further benchmark two pretrained sequence-to-sequence (Seq2Seq) models: BART~\cite{lewis-etal-2020-bart} and DistillBART~\cite{distill-bart}. We chose these two models as they achieve strong results on various summarization datasets. For BART, we train it from scratch, and for DistillBART, we use a version that has been fine-tuned on both XSum~\cite{narayan-etal-2018-dont} and CNN/DM~\cite{cnn-dataset} datasets. We use the fine-tuned DistillBART model in two settings: 1) further fine-tuning on \datasetname{}; 2) zero-shot setting. We use HuggingFace's Transformers to fine-tune both of these models.

The input to the Seq2Seq models is the concatenation of all tweets that are part of the timeline. Since each timeline in our dataset is annotated by three workers (\S\ref{sec:annotation}), it includes three summaries. To adapt the timeline summarization task to this setting, we pick the two summaries written by the two workers who agree the most based on the timeline extraction labels they assign to the tweets in each timeline. This reduces the variance between the summaries regarding their coverage of the crisis event described in the tweets that are part of the timeline.
During fine-tuning, we double the examples in {\train} by repeating each timeline twice, once for each summary. 
We describe all training settings and hyperparameters in Appendix \ref{sec:experimental-setup}.

\section{Results}
\subsection{Timeline Extraction}

Table~\ref{tab:te-dev} presents timeline extraction results on the {\dev} set. For evaluation, we use the average timeline-level accuracy. The naive majority class baseline marks every tweet as not part of the timeline and achieves 48.57 average timeline accuracy. This is expected since 10 timelines in the {\dev} set only contain tweets that are part of the timeline (\S\ref{sec:datastats}). The BERT-based sequence classification model achieves 74.64 average timeline accuracy, beating the BERT-GRU model, which achieves 65.86. Although the BERT-based sequence classification model has a limitation when extracting long timelines due to its limited size of 512 positional embeddings, it performs better than the BERT-GRU sequence labeling model. We attribute this to: 1) the careful preprocessing we did when constructing the sequences used to train the BERT-based sequence classification model, and 2) the limited data size, which might not enable the BERT-GRU model to fully capture the sequential relationship across the tweets in the timelines. 

For timeline extraction results on the {\test} set, we compare both the human-level performance and the best model's performance against experts' annotations.\footnote{It is worth noting that this is the same set that was used to estimate the annotation quality (\S\ref{sec:data-analysis})} To get the human-level performance on the {\test} set, we average the performance of the two workers who agree most across all timelines in {\test}. The best timeline extraction model achieves \textbf{73.51} average timeline accuracy, whereas the human-level performance is \textbf{88.98}. This highlights the difficulty of the timeline extraction task and indicates that more involved models are needed to close the gap between model performance and human-level performance. 

% \ke{Describe the results of timeline extraction accuracy vs length in Table 5 here}

\subsection{Timeline Summarization}
\label{sec:timeline-sum}
We evaluate all timeline summarization models with ROUGE \cite{lin-och-2004-automatic}. The summarization output of each model is evaluated in a multi-reference setting against the two summaries written by the two workers who agree the most based on the timeline extraction labels.
% they assign to the tweets in each timeline. 
We use SacreROUGE  \cite{deutsch-roth-2020-sacrerouge} to compute multi-reference ROUGE scores.

Table~\ref{tab:summarization-baselines} presents the F1 scores of ROUGE-1 (R1), ROUGE-2 (R2), and ROUGE-L (RL) of the timeline summarization models on the {\dev} set. The three naive baselines (first tweet, last tweet, and random tweet) achieve comparable performances, with R1 being the highest compared to R2 and RL. This is expected since the tweets in the timeline and the reference summaries (also tweets) all describe the same event, which leads to a considerable unigram overlap (i.e., R1).

For the Seq2Seq pretrained models, we present results in two settings: 1) using the oracle timeline; and 2) using the extracted timeline from the best timeline extraction model. In the oracle setting, we use the gold labels in the {\dev} set to identify tweets that are part of each timeline to the summarization models. The results of this oracle experiment estimate an upper bound for the summarization models used in this task. As shown in Table~\ref{tab:summarization-baselines}, the fine-tuned BART model achieves the best performance in terms of R1, R2 and RL. When we apply the best timeline extraction models to identify the tweets that are part of the timeline, we observe consistent conclusions. On {\test} set, the best summarization model (i.e., BART) achieves \textbf{47.05}, \textbf{25.40}, and \textbf{35.90} in R1, R2, and RL, respectively. 
% \ke{describe the ROUGE vs length error analysis in Table 5 here}
		
\begin{table}[t!]
\begin{center}
    
\begin{tabular}{|l|c|}
\cline{2-2}
\multicolumn{1}{l|}{} & \textbf{Accuracy} \\\hline
Majority Class & 48.57\\\hline
BERT & \bf 74.64\\\hline
BERT-GRU & 65.86  \\\hline
\end{tabular}
\end{center}

\caption{Results of timeline extraction models on {\dev}.}
\label{tab:te-dev}
\end{table}

% \begin{table}[t!]
% \begin{center}
    
% \begin{tabular}{lcc}
% \multicolumn{1}{l}{} & \textbf{Accuracy} & \textbf{F1} \\\hline
% Majority Class & 47.43 & 32.26 \\\hline
% BERT & \bf 75.09 & \bf 68.81 \\\hline
% BERT-GRU & 68.42 & 59.96 \\\hline

% \end{tabular}
% \end{center}

% \caption{Results of timeline extraction models on {\dev}.}
% \label{tab:te-dev}
% \end{table}

% \begin{table}[t!]
% \begin{center}
    
% \begin{tabular}{lcc}
% \multicolumn{1}{l}{} & \textbf{Accuracy} & \textbf{F1} \\\hline

% BERT & 72.76 & XX \\\hline
% Human-Level & \bf 89.34 & \bf XX \\\hline

% \end{tabular}
% \end{center}

% \caption{Results of the human-level and the best timeline extraction model performance on {\test}.}
% \label{tab:te-test}
% \end{table}

\begin{table}[t!]

\begin{center}
    
\setlength{\tabcolsep}{4pt} \scalebox{1}{
\begin{tabular}{|ll|c|c|c|}
\cline{3-5}
\multicolumn{2}{c|}{} & \bf R1 & \bf R2 & \bf RL \\\hline

First tweet & w/o TE & 32.15 & 14.35 & 24.09 \\
Last tweet & w/o TE & 28.69 & 12.03 & 21.65 \\
Random tweet & w/o TE & 30.56 & 13.61 & 23.24 \\\hline

\multirow{2}{*}{BART}  & Oracle & \bf 47.72 & \bf 26.16 & \bf 36.90 \\
                       & TE & \bf 45.32 & \bf 24.61 & \bf 34.68 \\\hline

DistillBART  & Oracle & 44.87 & 24.37 & 32.00\\
(Zero-shot)  & TE & 42.97 & 21.82 & 29.49 \\\hline

\multirow{2}{*}{DistillBART}  & Oracle &  47.50 & 25.11 & 34.64 \\
                              & TE & 45.04 & 22.61 & 32.51\\\hline
                        
\end{tabular}
}
\end{center}

\caption{Results of different timeline summarization models after timeline extraction and in oracle settings on {\dev}.}
\label{tab:summarization-baselines}
\end{table}

\begin{table*}[h!]
\centering

% \begin{center}
    
\setlength{\tabcolsep}{3pt} 
\scalebox{1}{
% \begin{tabular}{l @{\hskip 0.1in} cccc @{\hskip 0.3in} cccc}
\begin{tabular}{l @{\hskip 0.1in} |cccc ||  cccc|}
\cline{2-9}
% \cline{3-5}
 & \multicolumn{4}{|c||}{[All]} & \multicolumn{4}{c|}{[6~--]} \\\cline{2-9}
 & Acc & Coherence & Cov & Overall & Acc & Coherence & Cov & Overall \\
 % & \multicolumn{2}{c}{Accuracy} & \multicolumn{2}{c}{Coherence} & \multicolumn{2}{c}{Coverage} & \multicolumn{2}{c}{Overall} \\
 % & [1-5] & [6-] & [1-5] & [6-] & [1-5] & [6-] & [1-5] & [6-] \\
% \\\hline
\hline

% \multicolumn{1}{|l|}{Human} & \bf{4.76} & 4.85 & 4.17 & ~~\bf{4.15}$^\star$ & ~~\bf{4.73}$^\star$ & \bf{4.83}	& ~~\bf 3.97$^\star$	& ~~\bf 3.98$^\star$ \\
\multicolumn{1}{|l|}{Human} & \bf 4.75&	4.85&	4.14&	~~\bf{4.12}$^\star$ & ~~\bf{4.73}$^\star$ & \bf{4.83}	& ~~\bf 3.97$^\star$	& ~~\bf 3.98$^\star$ \\

\hline 

% \multicolumn{1}{|l|}{BART}  & 4.73 & \bf{4.87} &	4.00 &	4.01 & 4.59 &	4.81 &	3.43&	3.50 \\
\multicolumn{1}{|l|}{BART}  & 4.71 & \bf{4.86} &	3.91 &	3.93 & 4.59 &	4.81 &	3.43&	3.50 \\
\hline 
% \multicolumn{1}{|l|}{DistillBART}  &  4.28	&4.67	& \bf{4.24}	&3.93 & 4.56	& 4.70 &3.82	&3.74 \\
\multicolumn{1}{|l|}{DistillBART}  &  4.33	&4.68	& \bf{4.17}	&3.90 & 4.56	& 4.70 &3.82	&3.74 \\
\hline
% \multicolumn{1}{|l|}{DistillBART (Zero-shot)}  & 4.59	& 4.55	& 4.20	& 3.95 & 4.67&	\bf{4.83} &	 3.75 &	3.7\\
\multicolumn{1}{|l|}{DistillBART (Zero-shot)}  & 4.60	& 4.60	& 4.13	& 3.92 & 4.67&	\bf{4.83} &	 3.75 &	3.7\\

\hline 
% \multicolumn{1}{|l|}{Random Tweet}  & 4.76 &	4.73 &	3.53 &	3.52 & 4.53 &	4.64 & 2.72 &	2.79\\
\multicolumn{1}{|l|}{Random Tweet}  & 4.72 &	4.71 &	3.40 &	3.40 & 4.53 &	4.64 & 2.72 &	2.79\\
\hline 
% \multicolumn{1}{|l|}{Random Summary}  & 1.14 &	4.57 &	1.15 &	1.16 & 1.24	& 4.45	&1.25 &	1.25\\ 
\multicolumn{1}{|l|}{Random Summary}  & 1.16 &	4.55 &	1.16 &	1.17 & 1.24	& 4.45	&1.25 &	1.25\\ 
\hline
\end{tabular}
}
\caption{Results of human evaluation on human and model generated summaries. [All] means all timelines in the \test and [6~--] means ones with length $\ge 6$. $^\star$ denotes human summary is significantly (p < 0.05) better or worse than the best performing model based on one-sided Mann–Whitney U test~\cite{u-test}. Bold numbers denote best systems.}
\label{tab:summary_human_evaluation}
\vspace{-4mm}
\end{table*}

\paragraph{Human Evaluation}
We further conducted a human evaluation to better assess the quality of the summaries.  We take the whole {\test} set and we evaluate eight summaries per timeline: 1) the three human-written summaries; 2) three model-generated summaries using the three models we develop for timeline summarization in the oracle setting;  
% of workers who agree most based on the timeline extraction labels; 
3) two summaries from random systems: a random tweet that is part of the timeline, and a summary from a randomly selected timeline that belongs to a different crisis domain. The random systems add naive baselines and also serve as additional check on the annotation quality. 
% and 
% 4) three model-generated summaries using the three models we develop for timeline summarization in the oracle settings.
Following a similar process in \ref{sec:annotation}, we recruit a group of workers from MTurk with the same Location restriction~(QC1), who can pass our pilot study as a qualification test. 
% \joel{just a note that we spend a lot of time talking about recruitment and QC of the original turkers; how were these turkers here selected?  Should include a sentence} 
Each summary was evaluated by five different workers on a scale from 1 to 5 across four axes: Coherence, Accuracy, Coverage, and Overall quality, as was done by~\newcite{lai-etal-2022-exploration}. 
% Appendix \ref{sec:annotation_interface} describes the details about the annotation process.
Detailed instructions and annotation workflows are presented as
Figures \ref{fig:QE_task_instruction_page0}--\ref{fig:QE_task_rating_page3} in Appendix \ref{sec:annotation_interface}.

% \ke{describe the QE results.}

Table \ref{tab:summary_human_evaluation} presents the results. Looking at the results over all the timelines, the human-written summaries are significantly better than the ones generated by models, in terms of overall quality with an average rating of $4.12$. The human-written summaries are also better in terms of accuracy, however, this was not statistically significant compared to the average accuracy ratings assigned to the models generated summaries. The models generated summaries have higher average ratings in terms of coherence and coverage, but this is not statistically significant compared to the human-written summaries. Notably, the gap of the ratings becomes larger for longer timelines (i.e., timelines with 6 or more tweets). The human-written summaries are significantly better in terms of accuracy, coverage, and overall. In terms of coherence, they are on par with the models generated summaries. This highlights the shortcomings of the summarization models when it comes to long timelines.

% As we can see, human summaries consistently show better overall quality compared to summaries generated by our strong baseline models (e.g., BART). 
% The gap becomes larger and significant across other measures as well (i.e., accuracy and coverage), for longer timelines. 

% \ke{show an example for model outputs to highlight the challenges of the summarization task: duplication, hallucination, the order of the events appear in the timeline which is one type of hullucination}

% \ke{Note: the screenshot of QE rating scale in the Appendix is 1-7, which should be updated. We actually use 1-5 for the latest QE}

% \subsection{Error Analysis} \ke{to remove}
% We conduct an error analysis over the outputs of the best models for timeline extraction and summarization on {\dev}. For timeline extraction, 69 timelines~(80.2\%) have at least one error, in which 253 out of 962 tweets~(26.3\%) are wrong.

% \ke{not sure where to put \ref{tab:error_analysis}}
\subsection{Error Analysis}
We conduct an error analysis over the outputs of the best models for timeline extraction and summarization on {\dev}. For timeline extraction, 66 timelines~(77\%) have at least one error, in which 272 out of 958 tweets~(28\%) are wrong.

\begin{table}[ht!]
\centering
\setlength{\tabcolsep}{3pt}
\begin{tabular}{|l|c|c|c|c|}
\cline{2-5}
\multicolumn{1}{}{} & \multicolumn{1}{|l|}{Extraction}  & \multicolumn{3}{|c|}{Summarization} \\\hline

\textbf{Length} & \textbf{Accuracy} & \textbf{R1}& \textbf{R2}& \textbf{RL}\\\hline
[1--5]  & 76.97 & 49.43 & 29.48 & 39.25  \\\hline
[6--12] & 74.24 & 43.92 & 23.19 & 33.62  \\\hline
[13--25] & 73.59 & 43.71 & 22.93 & 32.65  \\\hline
[26--] & 67.37 & 35.10 & 11.87 & 21.73  \\\hline

\end{tabular}
\caption{Best timeline extraction and summarization models performances on {\dev} based on timelines' lengths.}
%across the four target corpora}
\label{tab:error_analysis}
\vspace{-1mm}
\end{table}

Table~\ref{tab:error_analysis} presents timeline extraction and summarization results of the best models based on different timeline lengths over the {\dev} set. We observe that the model performance decreases with increasing timeline length for both timeline extraction and summarization tasks, with significant drops in accuracy and ROUGE scores when timelines get longer.  Moreover, we inspected some of the generated summaries manually and we noticed that most of the summarization errors were due to hallucinations or to copying specific sentences that are present in the timeline without covering all the important event details described in the timeline. For instance, for the timeline in Figure~\ref{fig:main-example}, the best baseline summarization model (i.e., BART) generated the following summary: ``\textit{A large fire burning in northeast Fresno near Woodward Lake sent plumes of smoke into the air above the city. Officials say the fire started as a commercial fire}''. The two sentences in the generated summary are copied verbatim from the tweets in the timeline. This highlights the need for better models to capture the important details mentioned in the timeline.

\section{Conclusion}
We presented \datasetname{}, the first dataset on local crisis timelines extracted from Twitter and the first to provide human-written summaries on information extracted from Twitter.
We showed that \datasetname{} supports two downstream tasks: Timeline Extraction and Timeline Summarization. Our experiments with SOTA baselines indicate that both of these tasks are challenging and encourage future research. Our dataset further provides a resource for developing methods on utilizing micro-blogs toward aiding first-responders in evaluating ongoing crisis events.
In future, we plan to explore models that can solve both tasks in a joint setting by extracting new information from each update point and then summarizing those.
This dataset can also be expanded by additional annotation to enable abstractive entity-based understanding of the event flow~\cite{mishra2018tracking,faghihi2021time}. 
% Additionally, this dataset can facilitate research on providing time-sensitive information for first responders during crisis events by enabling real-time crisis detection and tracking. 
% As a future direction, we plan to provide additional annotation on event timelines which can enable abstractive entity-based understanding of their flow.

% \joel{can we make sure that the AI for Social Good element is reinforced here}

\section*{Acknowledgements}

We thank Aoife Cahill for providing detailed feedback on this paper,
Alison Smith-Renner and Wenjuan Zhang for helpful discussions on MTurk annotations, our colleagues at Dataminr, and the
\textsc{emnlp} reviewers for their helpful and constructive comments.

% \newpage

\section*{Limitations}
This study has some limitations on the dataset generation workflow.
First, our noisy timeline collection process is not a comprehensive and exhaustive extraction to find all available information about a local crisis event. Here, our focus is to provide a representative dataset rather performing a comprehensive set containing all the local crisis events and their updates.
Second, the proposed noisy timeline collection pipeline is highly dependent to the performance of the entity extraction modules, especially for the location extraction, and the accuracy of the OpenStreetMap API to find the correct physical address of each location mention. Accordingly, replicating the same process for other language or based on other locations may be hard because of such dependencies. Furthermore, the online clustering method used in this paper has a set of hyperparameters which are tuned heuristically from a small set of experiments. More comprehensive and large scale experiments on tuning those parameters could potentially impact the quality of the generated timelines.
Generating similar results by following our noisy timeline collection process is in general limited by the users' access to the public Twitter stream and the changes in the available posts~(they may become restricted or deleted).

% In terms of making the dataset publicly available, we are limited to sharing only the tweet ids of the generated timelines which may require user access to the public Twitter API for recovering the original text. Over time, some tweets may get removed or deleted, we recommend to use the historical snapshot of Twitter to recover such lost information.
% In our final crisis timelines, there may be mentions of people, or organization names. However, as we do not share the text containing those information, people can only access those using the publicly available information.
% Additionally, we have used the replication factor 3 to gather annotation for each of the timelines in \datasetname{}. A higher replication factor can indeed improve the quality of the final labels, especially on subjective decisions. 

% \section*{Acknowledgement}

\bibliography{anthology,acl2020, custom}
\bibliographystyle{acl_natbib}

\appendix
\section{Noisy Timeline Collection}
\label{sec:app_collection}
In this section, we provide some additional information on the noisy timeline collection process.
\subsection{Location and Date Selection}
We build \datasetname{} by limiting the search to four crisis domains: wildfire, local fire, traffic, and storm. We collect \datasetname{} across seven states in the USA and two states of Australia during 2020 and 2021.
The selected states in the USA are New York, Illinois, Massachusetts, California, Texas, and Florida. From Australia, we have selected the states of New South Wales and Victoria. These states are selected as they are some of the areas subject to the most events falling into our crisis domains of interest. California, Victoria, and New South Wales are mainly selected due to the abundance of wildfires in hot seasons. Texas and Florida have been selected as they are both subject to wildfires and storm events during different months. Massachusetts and Illinois are selected first because those areas are frequently subjected to bad weather, and second as they contain big cities subject to traffic and local fire events alongside New York. 
\subsection{Keyword Filtering}
\label{sec:app_keyword}
This step selects the subset of filtered tweets related to a specific crisis category of interest. Ideally, this task can be performed by a neural model trained on a large set of tweets with labeled data indicating the categories. However, as such a large amount of labeled data is unavailable, we rely on a common approach of designing keywords. We carefully curate a keyword list for each of our categories of interest by employing expert knowledge gathered through over-viewing crisis stories in news, social media, and related big disaster events of the past. Table \ref{tab:keywords_example} 
shows some example keywords used in each of the crisis domains in \datasetname{}. Please note that these lists are generic for each category and not specific to unique events. For instance, instead of defining keywords specific to an event called ``Hurricane Ida'', our keywords list includes phrases such as ``fallen tree'',  ``building collapse'', or ``storm''. 
The quality of the keywords list is crucial to the final quality of the generated timelines, and we polish this list multiple times based on small sets of experiments before using them for the final task.
Each keyword can be either a single word or multiple words. To avoid making a long list of keywords and ensure that different lexical formats of the same word are still considered in the keyword matching process, we maintain both a lower-cased and a lemittized version of the keywords and the tweet's text. If any of the keywords exist in the text and in any of these formats, then the tweet is considered related to the category. The multi-word keywords are not considered n-grams but as an indication that all the words in the keyword should exist in the text, even if not as a sequence. This approach will not be comprehensive or exhaustive but rather representative of each crisis domain. Improving this method to be more encompassing is an area for future research.

% Figure \ref{fig:keywords} 
% Table \ref{tab:keywords_example} 
% shows some example keywords used in each of the crisis domains in \datasetname{}.
% \begin{figure*}[h]
%     \centering
%     \includegraphics[width=\linewidth]{figs/sample_keywords.png}
%     \caption{Sample keywords for each crisis domain in \datasetname{}}
%     \label{fig:keywords}
% \end{figure*}

\begin{table*}[ht!]
\centering
\setlength{\tabcolsep}{3pt}
\begin{tabular}{|l|l|}
% \cline{2-5}
\hline
Wildfire & wildfire, bushfire, campfire, volcano erupt, forest fire, ash fall, vegetation fire   \\\hline
Local fire & house fire, structure fire, building fire, gas leak, fire alarm, tower burned  \\\hline
Storm &  tornado, shelter, storm damage, building collapse, storm collapse, roof damage, fallen tree \\\hline
Traffic & traffic delay, road blocked, car fire, crash, injury,  rollover, accident, stalled vehicle  \\\hline

\end{tabular}
\caption{Sample keywords for each crisis domain in \datasetname{}.}
%across the four target corpora}
\label{tab:keywords_example}
\end{table*}

\subsection{Clustering Hyper-Parameters}
We use the following parameters shared among all different domains:
$s_{hashtag}$ is set to $0.2$ and  $s_{dist}$ is $0.3$. The $sim_{threshold}$ and $time_{threshold}$ are set to $0.7$ and $15$ hours respectively. The $time_{threshold}$ is reduced to $3$ hours to avoid merging various accidents at different times but in the same location. The $min_{dist}$ and $max_{dist}$ are set to $0.4$ kilometer and $4$ kilometer for fire and traffic events while having the larger range of $0.4$ and $10$ kilometers for the wildfire and storm extraction.
The $expiration_{threshold}$ is set to $15$ hours and $tweet_{threshold}$ is $4$.
\subsection{Automatic and Manual Cluster Merge}
The online clustering approach may fail to properly relate some tweets due to missing entities, clustering method hyper-parameters, or the difference in the description of various angles of the same story. To address this, we merge clusters by comparing all cluster heads with each other and combining those with a higher similarity score than a threshold $s_{head}$. Additionally, we use human feedback to merge clusters where their similarity score is below $s_{head}$ but higher than a second hyper parameter $s_{head}^{min}$. This process can compensate for a portion of the missing entities due to entity extraction accuracy or tweets' informal text.

\subsection{Duplicate Removal}
Here, the goal is to remove tweets from the same cluster with identical or similar text. The duplicate removal relies on a fuzzy string sequence-matching technique to compute the similarity between a pair of tweets. We simply go over each tweet sorted in chronological order and remove the ones that match any of the previous tweets in the same cluster with a matching score higher than $d_{match}$. 

\subsection{Noise Removal}
Although the keyword filtering step would reduce the number of tweets unrelated to the category of interest, this step aims to further remove the unrelated clusters generated from the pipeline. To do so, we use a neural model for detecting the informativeness of tweets' text by training it on the available labeled data~\cite{alam2021humaid}. As the definition of informativeness in our domain differs from the data source, we further define a mapping between their label set and the informativeness as we refer to it. Hence, any fine-grained label related to personal emotions, prayers, or donations is removed from the informative set.
Since this process is very category relevant and the existing labeled datasets do not cover all of our categories of interest, we can only apply this step to those categories where a representative subset exists for them in the available resources. 
We formulate this task as a Boolean tweet classification task to predict the tweet informativeness by applying a linear classification module on top of the aggregation token of a transformer-based language model.

\section{Noisy Clusters Selection for Human Annotation}
\label{sec:cluster_selection}
This section gives more details on how we select noisy clusters for human annotation. Our goal is to select ~1000 clusters that contain enough tweets describing an evolving crisis event.
In particular, we first identify a ``seed tweet'' as the first observed post mentioning a local crisis event, then roughly check whether the following tweets in the cluster contain updates about the same event. Taking the example in Figure \ref{fig:main-example}, we first see ``a vegetation fire happened'' in the seed tweet, then see ``plumes of smoke'', ``windy condition impacts fire control''. 
We select clusters across four crisis domains, including wildfire, fire, storm, and traffic, depending on how frequently each type of event happens in extracted noisy clusters. 

\begin{table*}[th!]
\centering
\setlength{\tabcolsep}{2pt}

\begin{tabular}{|l|l|l|l|l||l|l|l|l||l|l|l|l|}
\cline{2-13}
\multicolumn{1}{l|}{} & \multicolumn{4}{|c||}{\bf Train} & \multicolumn{4}{|c||}{\bf Dev} & \multicolumn{4}{|c|}{\bf Test}\\\hline
\bf Domain & [1 -- 5] & [6--12] & [13--25] & [26 --] & [1 -- 5] & [6--12] & [13--25] & [26 --] & [1 -- 5] & [6--12] & [13--25] & [26 --] \\\hline\hline
\bf{Wildfire} & 125 & 101  & 42 & 35 &  14 & 11  & 5 & 4 &  36 & 28  & 12 & 10\\\hline
\bf{Traffic} & 113 & 66  & 15 & 9 &   13 & 8  & 2 & 2 &  32 & 21  & 3 & 3\\\hline
\bf{Fire} & 54 & 32  & 15 & 8 &   7 & 3  & 2 & 1 &  15 & 10  & 5 & 3\\\hline
\bf{Storm} & 14 & 30  & 18 & 12 &   2 & 4  & 3 & 2 &  4 & 2  & 0 & 0\\\hline
\bf{Other} & 11 & 5  & 0 & 1 &  2 & 1  & 0 & 0 &  5 & 9  & 6 & 4\\\hline

\end{tabular}

\caption{Aggregated data statistics based on crisis domains and timeline lengths of the {\train}, {\dev}, and {\test} splits.}
\label{tab:stats-splits}
\end{table*}

\subsection{Sample noisy timelines}
\begin{figure*}[h]
    \centering
    \includegraphics[width=.7\linewidth]{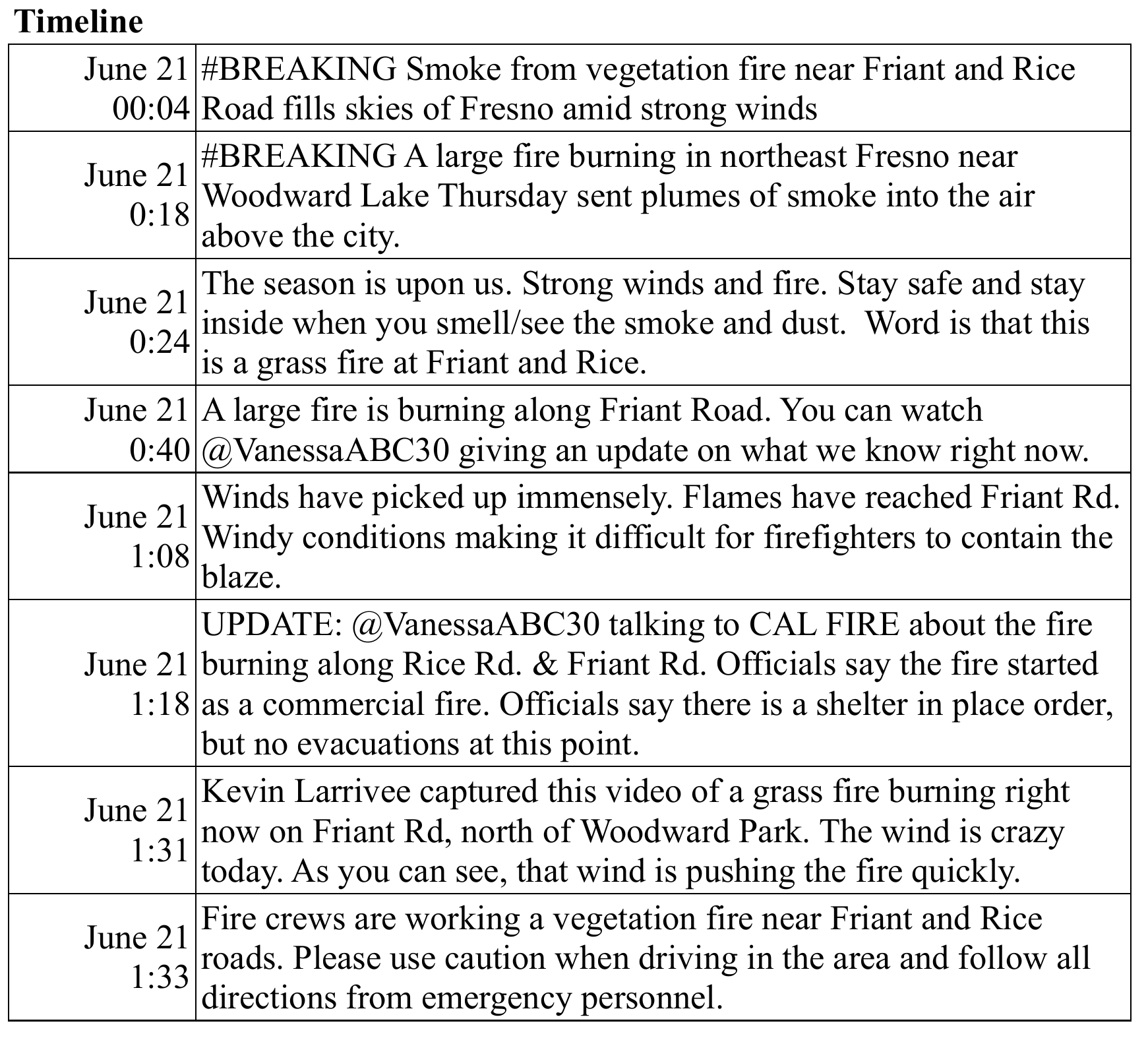}
    \caption{Sample noisy timeline of a wildfire event. This is the same example depicted in Figure \ref{fig:main-example}.}
    \label{fig:wildfire_sample}
\end{figure*}

\begin{figure*}[h]
    \centering
    \includegraphics[width=.7\linewidth]{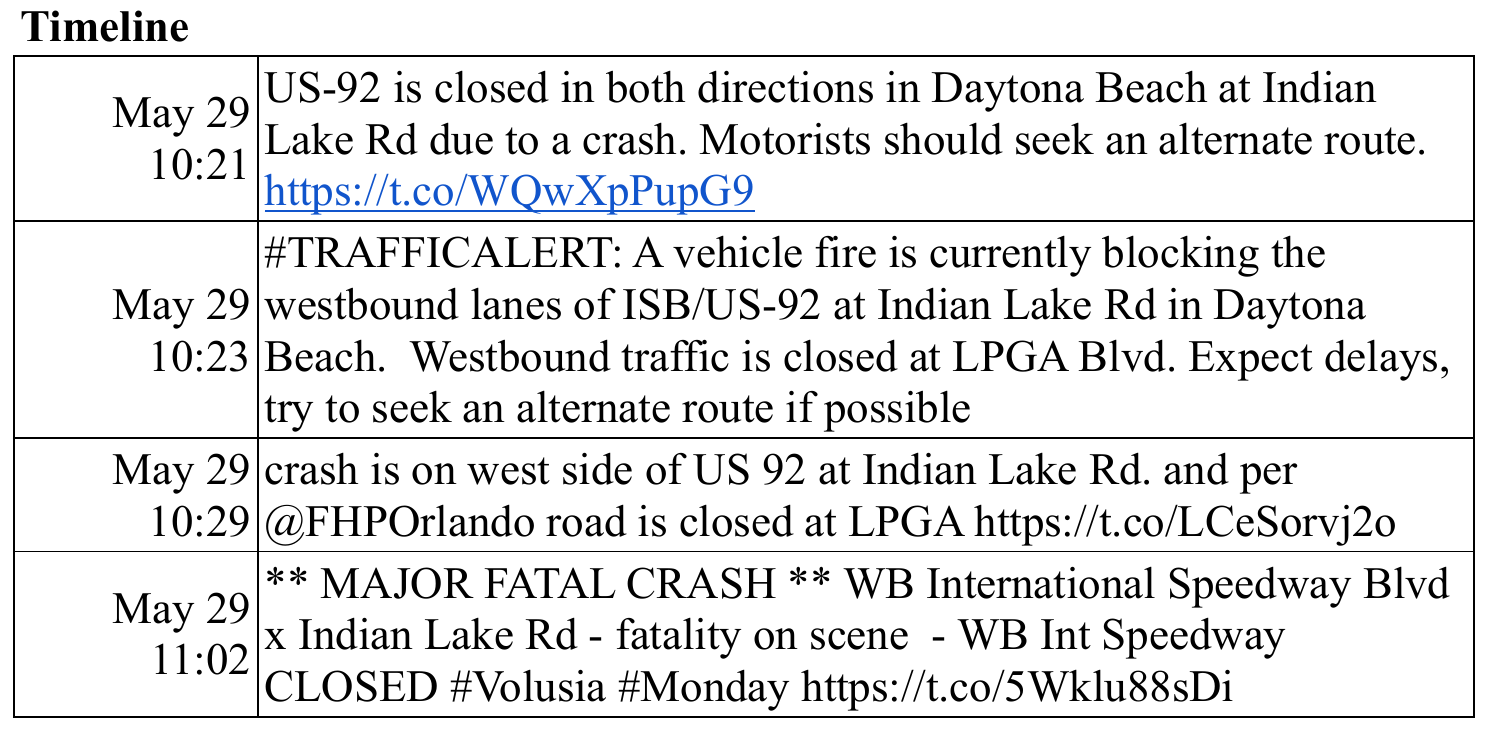}
    \caption{Sample noisy timeline of a traffic event.}
    \label{fig:traffic_sample}
\end{figure*}

\begin{figure*}[h]
    \centering
    \includegraphics[width=.7\linewidth]{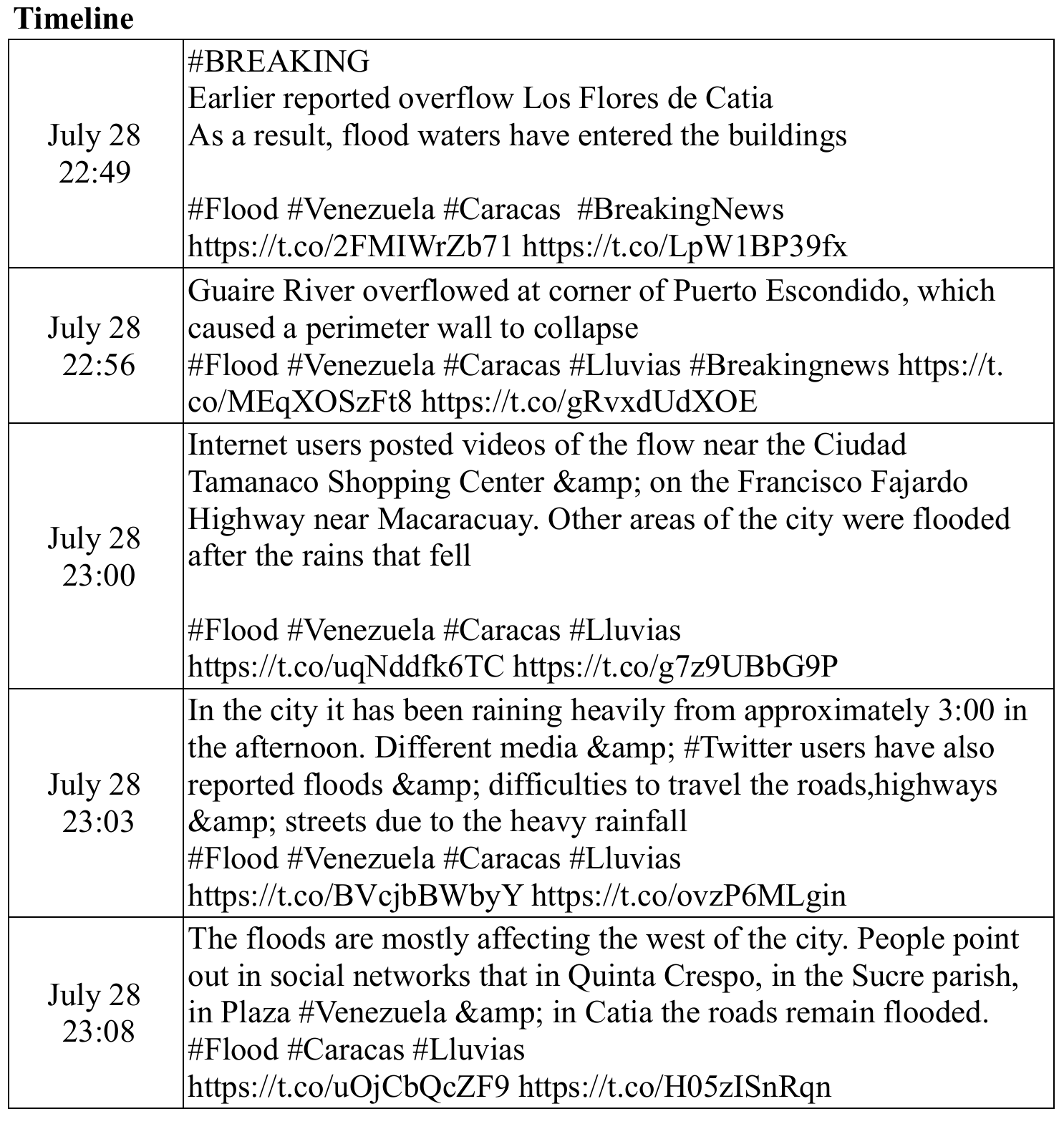}
    \caption{Sample noisy timeline of a storm event.}
    \label{fig:storm_sample}
\end{figure*}

\begin{figure*}[h]
    \centering
    \includegraphics[width=.7\linewidth]{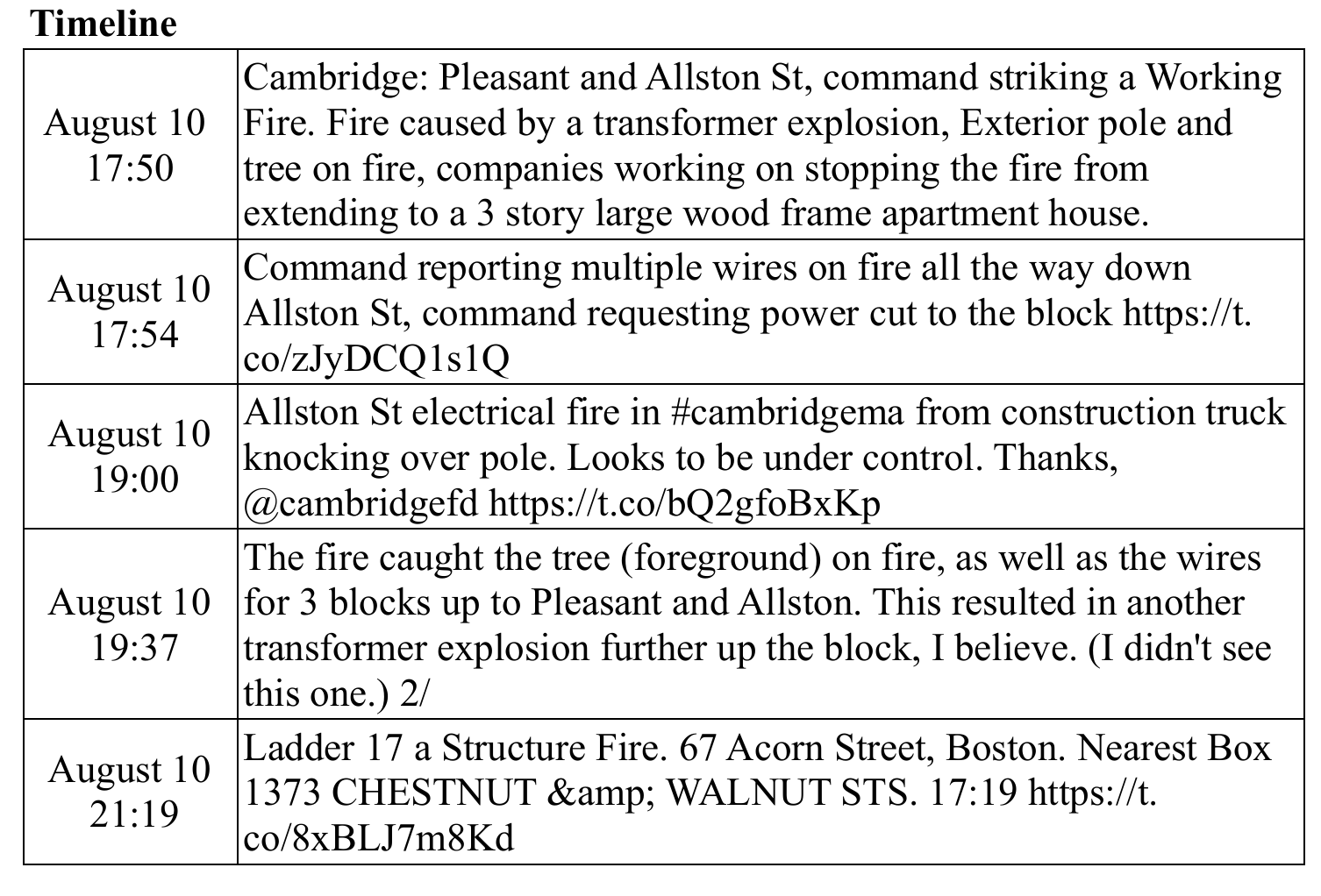}
    \caption{Sample noisy timeline of a fire event.}
    \label{fig:fire_sample}
\end{figure*}

Figure \ref{fig:wildfire_sample} shows the noisy timeline of the same annotated example from Figure \ref{fig:main-example}. Figure \ref{fig:traffic_sample}, \ref{fig:storm_sample}, and \ref{fig:fire_sample} show sample noisy timelines from other domains in \datasetname{}.

\section{Detailed Experimental Setup}
\label{sec:experimental-setup}
\subsection{Timeline Extraction}
\paragraph{BERT} We fine-tune BERT base uncased on a single GPU for 10 epochs with a learning rate of 5e-5, batch size of 32, a seed of 42, and a maximum sequence length of 512.
At the end of the fine-tuning, we pick
the best checkpoint is based on the performance on
the {\dev} set. 

\paragraph{BERT-GRU} For the BERT-GRU sequence labeling model, we use BERT base uncased to get the contextual representation for each tweet. The GRU has one layer with a hidden size of 128. The model was trained for 50 epochs with early stopping after five epochs if the performance did not improve on the {\dev} set. We use a learning rate of 5e-5, a batch size of 16, and a seed of 42.

\subsection{Timeline Summarization}
We fine-tune BART based on a single GPU for ten epochs with a learning rate of 5e-5, batch size of 16, a seed of 42, and a maximum target sequence length of 512. For DistillBART, we use the same hyperparameters. During inference, we use beam search with a beam size of 4. At the end of the fine-tuning, we pick
the best checkpoint is based on the performance on
the {\dev} set.

\section{Detailed Data Splits}
\label{sec:data-splits}
Detail statistics on {\train}, {\dev}, and {\test} are shown in Table~\ref{tab:stats-splits}.

% \section{Procedures}

% \paragraph{Timeline Extraction \& Summarization}

% \paragraph{Summarization Quality Estimation}

% \section{Annotation Workflow \& Quality Control}

% To ensure that we recruit high-quality workers for our annotation tasks, we adopt multiple Quality Control (QC) steps in our annotation protocol. First, we apply \textbf{location \& worker type restrictions (QC1)} to limit the worker pool to Master Workers in the United States only since we need English Native speakers to review the timeline in English and write summaries or rate a summary written in English. We then launch a \textbf{qualification task (QC2)} to test the worker's ability to perform this task. The qualification tasks consist of 5 timelines for the workers to extract and write summaries. The worker's performance on the qualification task is then \textbf{evaluated against the ground truth answers and reviewed by the authors (QC3)} to exclude the workers who performed poorly on the task. We use the same group of workers from the Timeline Extraction \& Summarization task for the Summarization Quality Estimation task because of their familiarity with it.

% \section{Compensation}

% We compensate the workers with a rate of $\$2.5$ per qualification task and $\$3.0$ per HIT for the Timeline Extraction \& Summarization task and $\$1.5$ per HIT for the Summarization QE task. Each task is followed with a one-time bonus that makes the final rate over $\$10$ per hour.

\section{Annotation Interface}
\label{sec:annotation_interface}
\begin{figure*}[h]
    \centering
    \includegraphics[width=.7\linewidth]{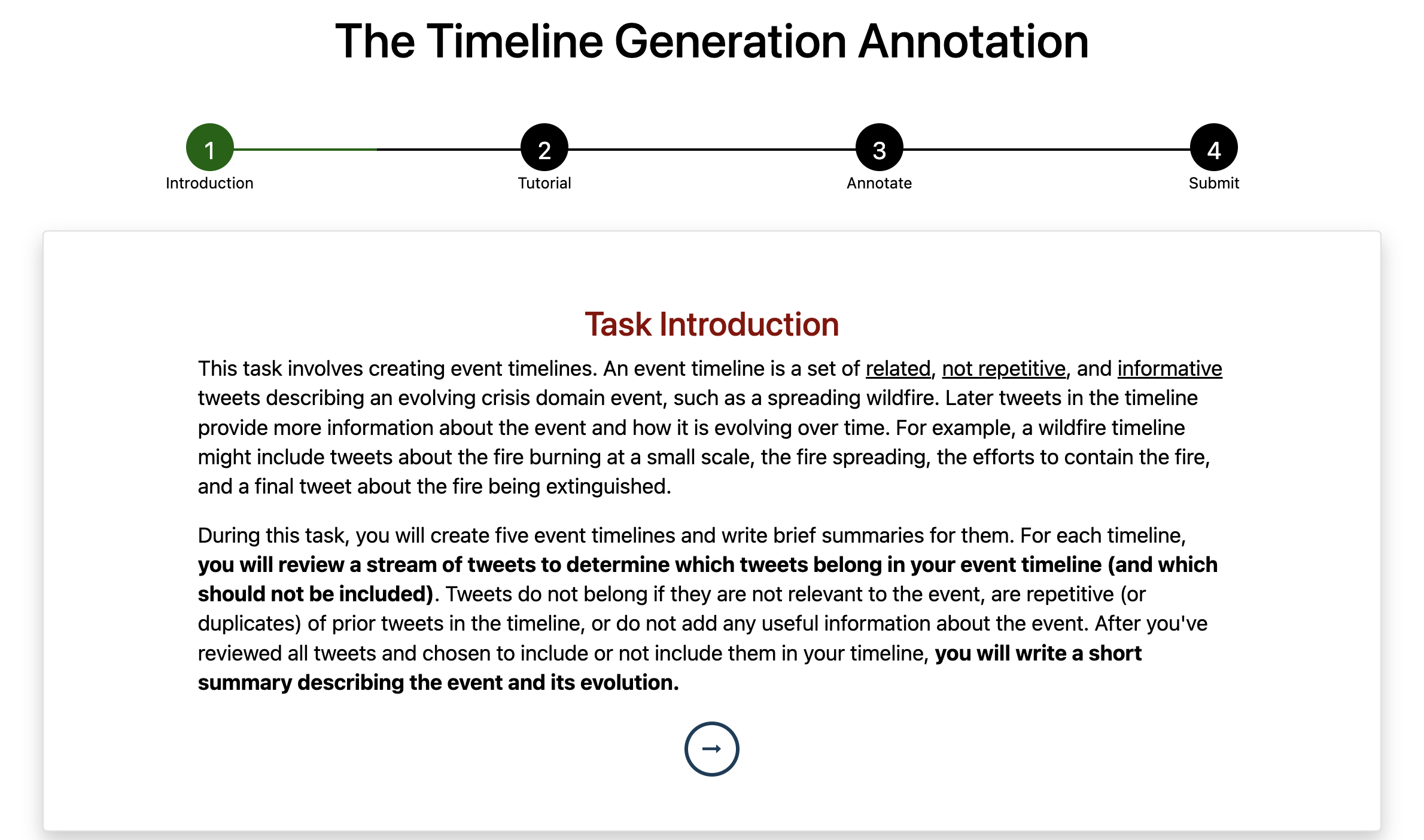}
    \caption{The first page of our annotation interface for the Timeline Extraction \& Summarization task, which contains the task introduction and a brief instruction.}
    \label{fig:TS_task_page1}
\end{figure*}

\begin{figure*}[t]
    \centering
    \includegraphics[width=.7\linewidth]{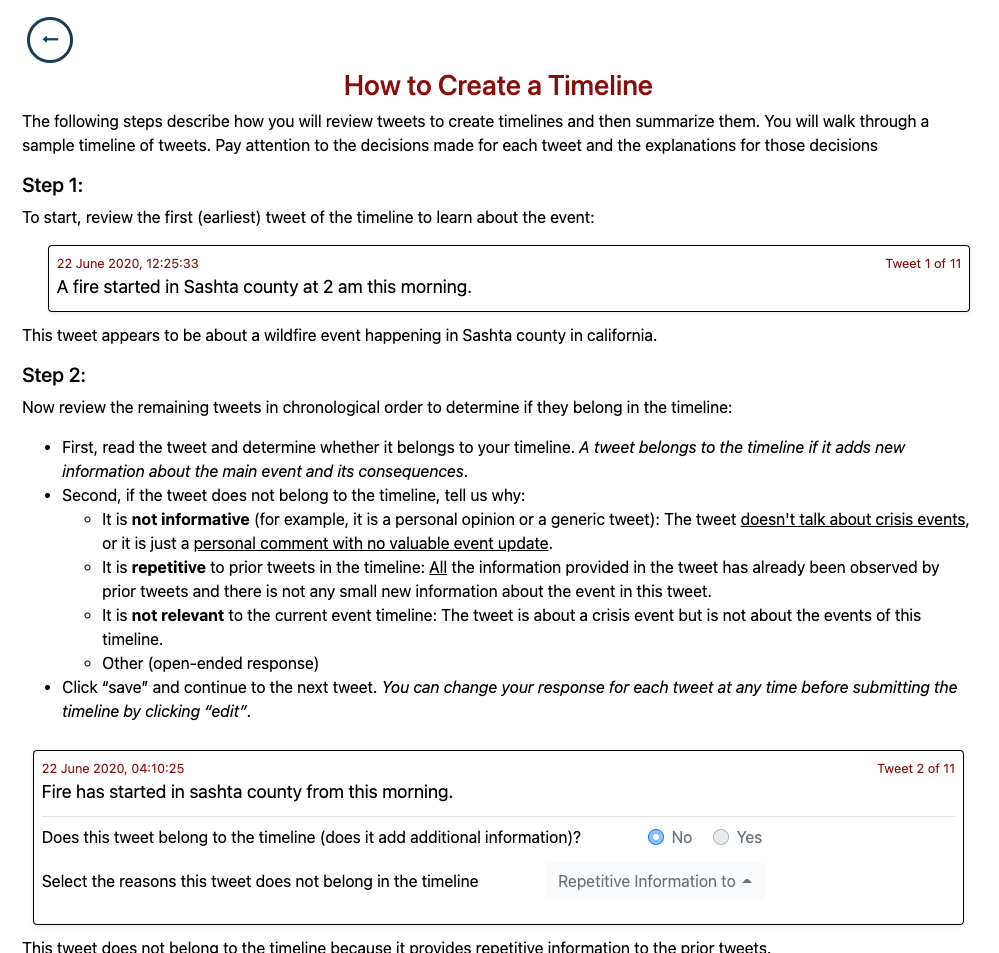}
    \caption{Step 1 - 2 of the annotation instructions in our annotation interface for the Timeline Extraction \& Summarization task}
    \label{fig:TS_task_page2_upper}
\end{figure*}

\begin{figure*}[t]
    \centering
    \includegraphics[width=.7\linewidth]{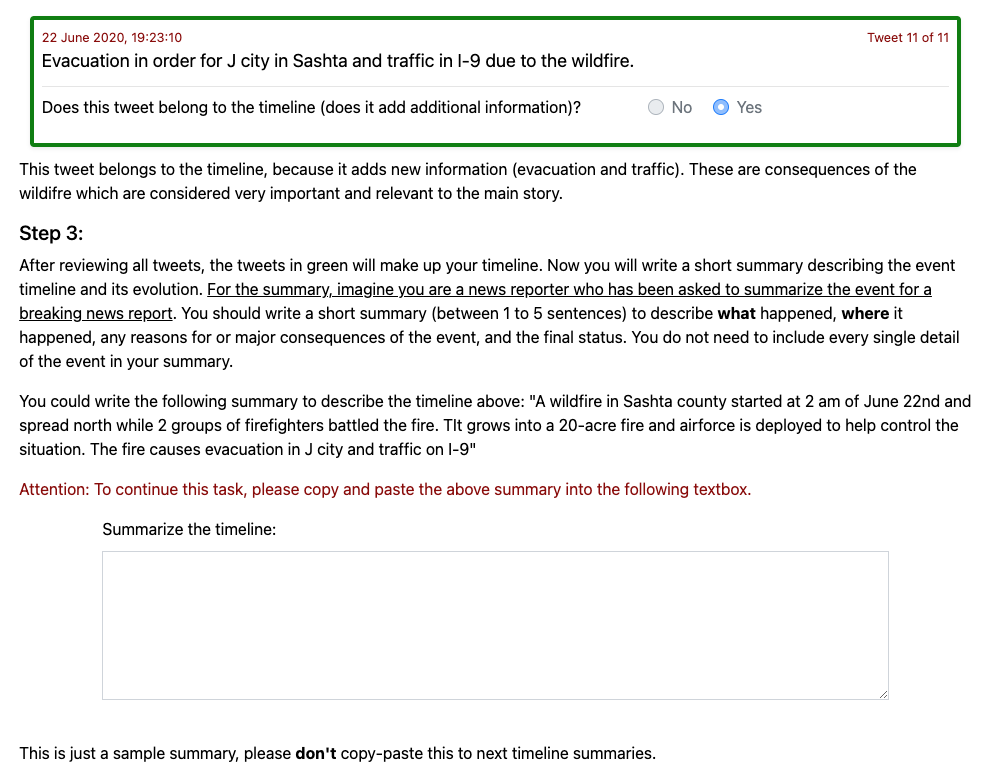}
    \caption{Step 3 of the annotation instructions in our annotation interface for the Timeline Extraction \& Summarization task}
    \label{fig:TS_task_page2_middle}
\end{figure*}

\begin{figure*}[t]
    \centering
    \includegraphics[width=.7\linewidth]{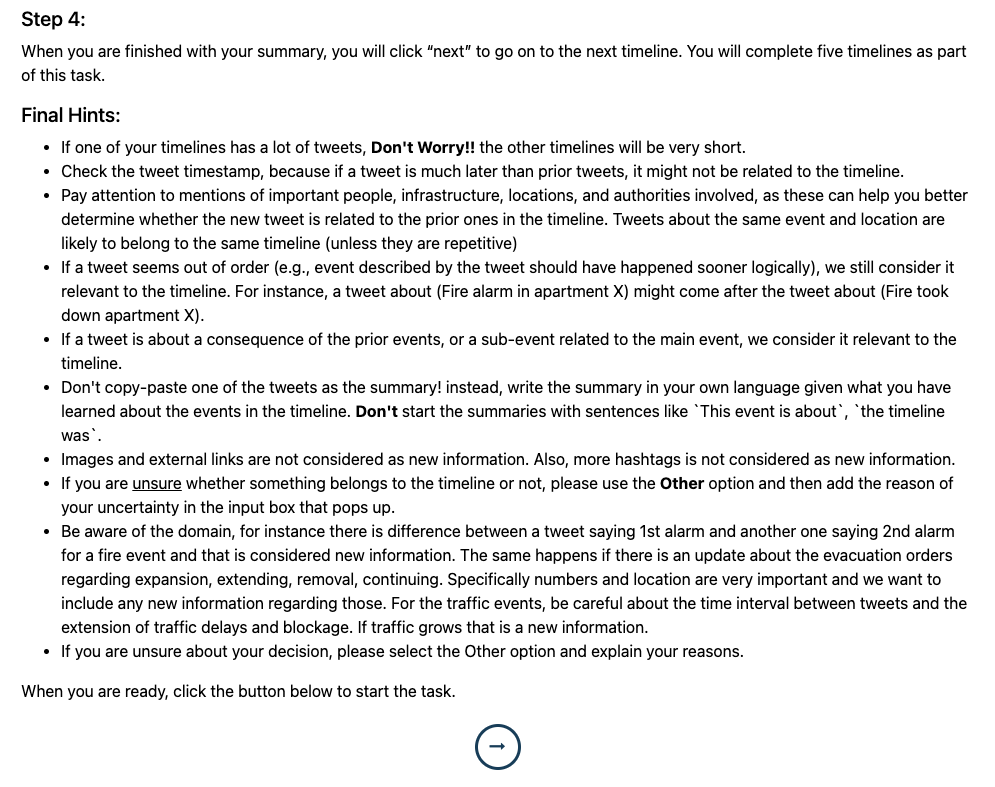}
    \caption{Step 4 and final hits of the annotation instructions in our annotation interface for the Timeline Extraction \& Summarization task}
    \label{fig:TS_task_page2_lower}
\end{figure*}

\begin{figure*}[t]
    \centering
    \includegraphics[width=.7\linewidth]{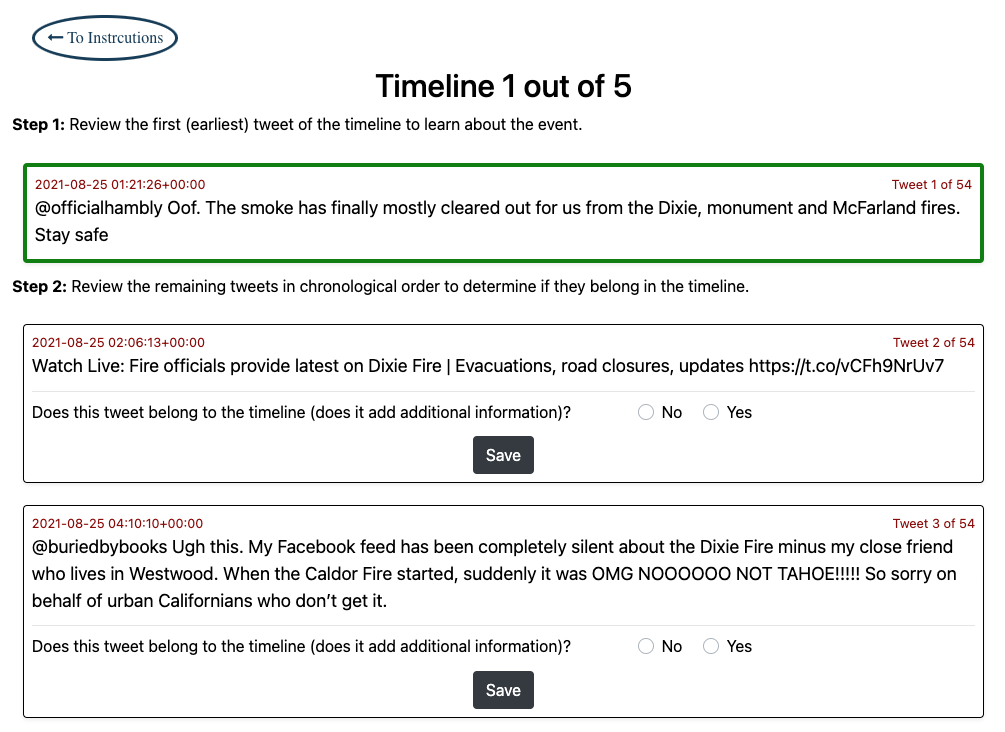}
    \caption{Step 1 - 2 of the task page in the annotation interface for the Timeline Extraction \& Summarization task}
    \label{fig:TS_task_page3_upper}
\end{figure*}

\begin{figure*}[t]
    \centering
    \includegraphics[width=.7\linewidth]{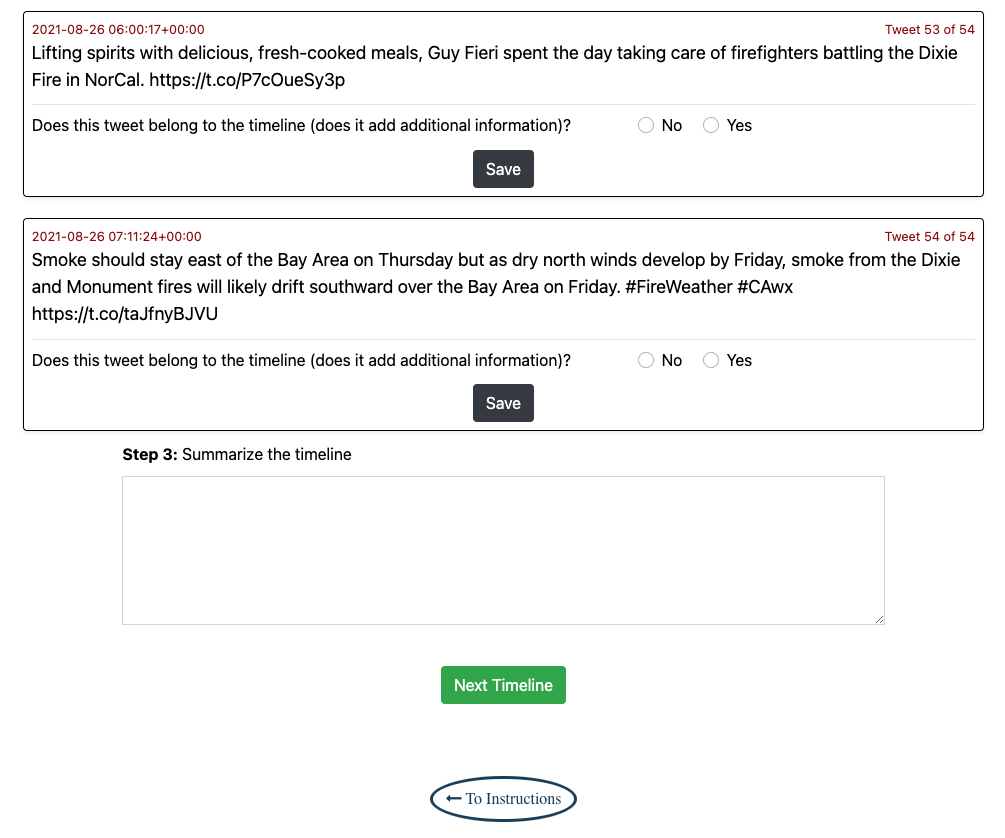}
    \caption{Step 3 of the task page in the annotation interface for the Timeline Extraction \& Summarization task}
    \label{fig:TS_task_page3_lower}
\end{figure*}

\begin{figure*}[t]
    \centering
    \includegraphics[width=.7\linewidth]{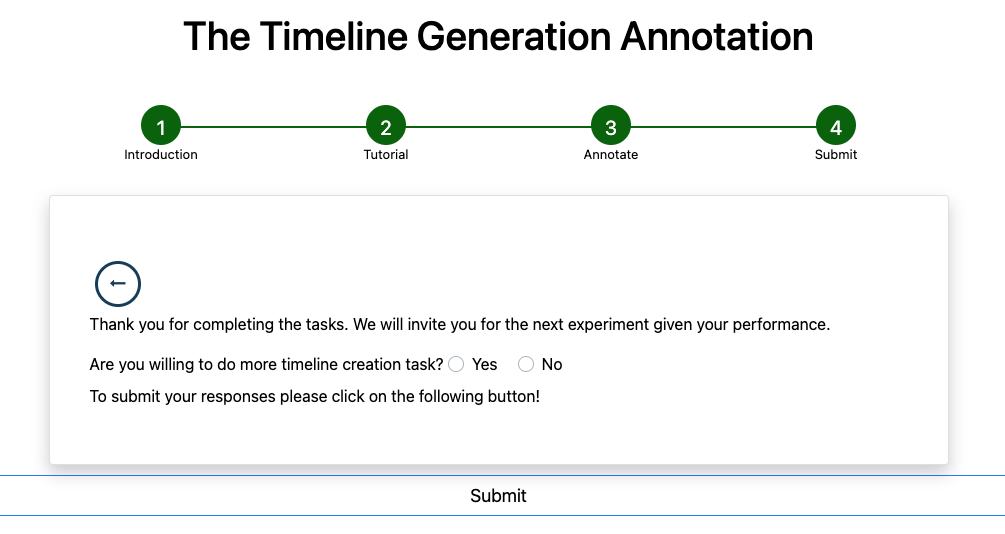}
    \caption{Completion page in the annotation interface for the Timeline Extraction \& Summarization task}
    \label{fig:TS_task_page4_all}
\end{figure*}

\begin{figure*}[t]
    \centering
    \includegraphics[width=.7\linewidth]{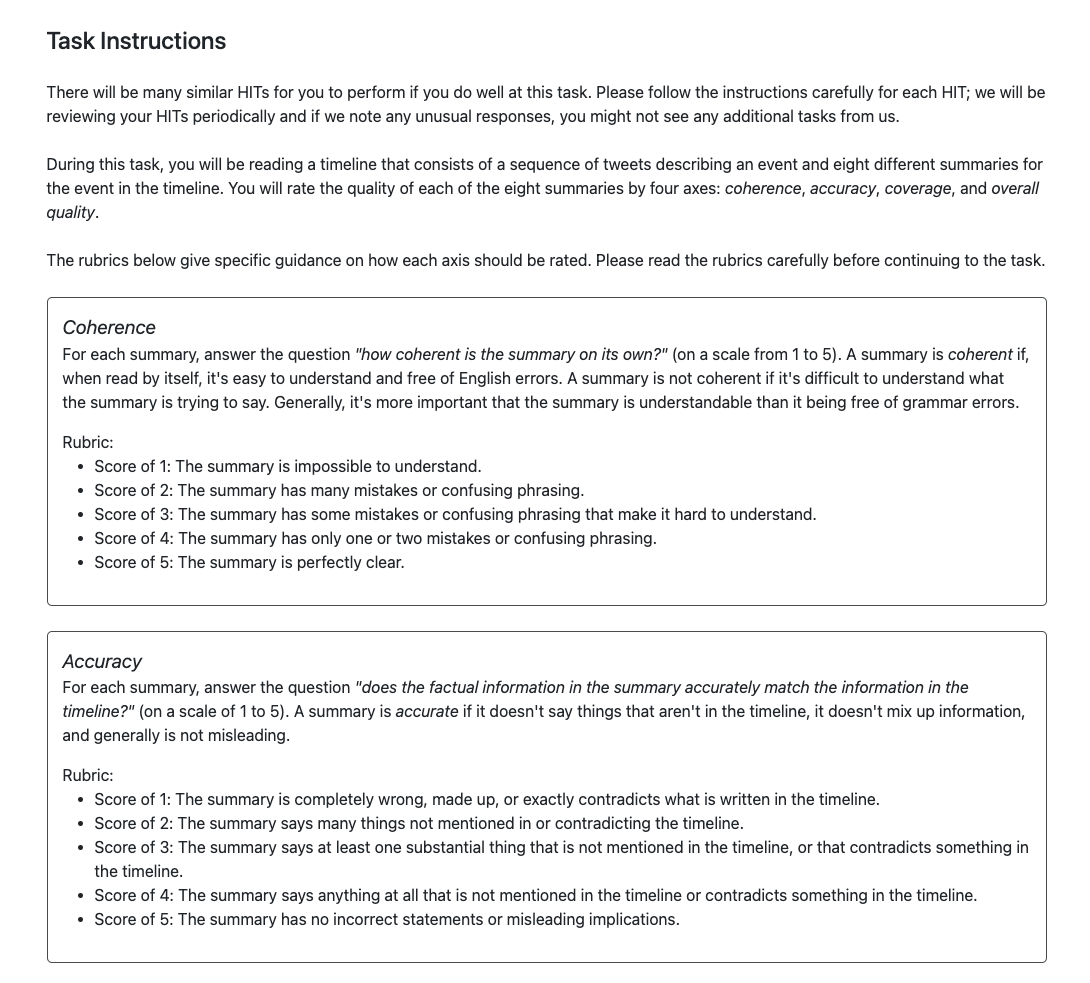}
    \caption{Instruction page of the Summarization Quality Estimation Task part 1.}
    \label{fig:QE_task_instruction_page0}
\end{figure*}

\begin{figure*}[t]
    \centering
    \includegraphics[width=.7\linewidth]{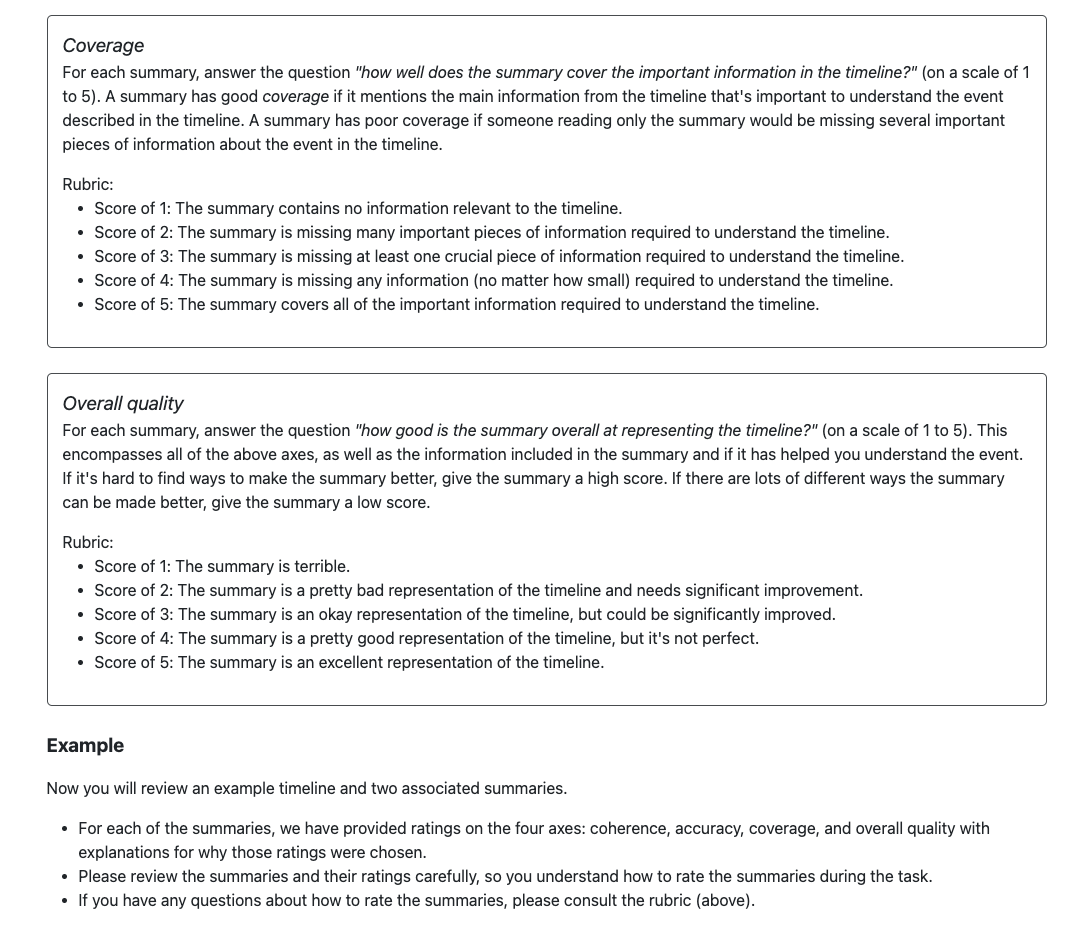}
    \caption{Instruction page of the Summarization Quality Estimation Task part 2.}
    \label{fig:QE_task_instruction_page1}
\end{figure*}

\begin{figure*}[t]
    \centering
    \includegraphics[width=.7\linewidth]{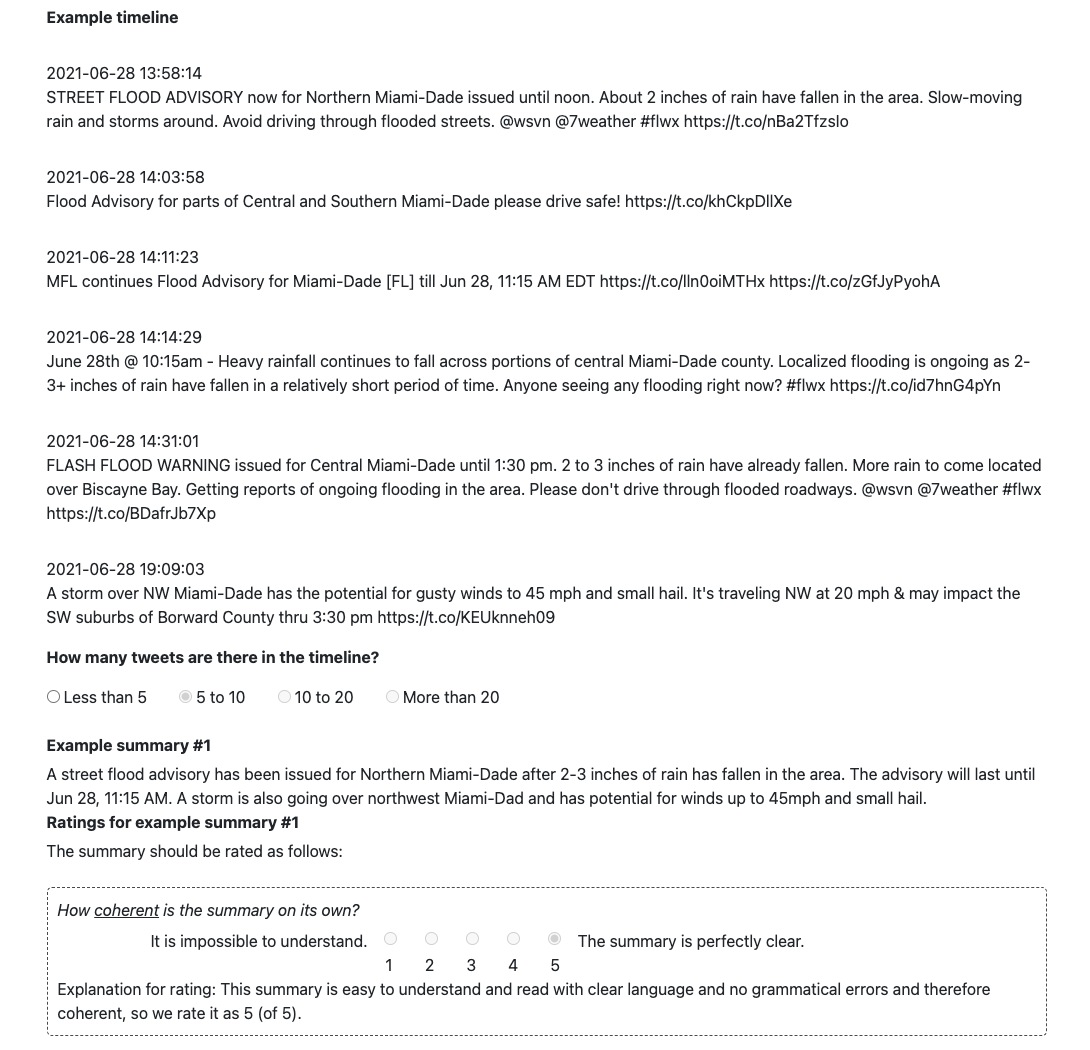}
    \caption{Instruction page of the Summarization Quality Estimation Task part 3.}
    \label{fig:QE_task_instruction_page2}
\end{figure*}

\begin{figure*}[t]
    \centering
    \includegraphics[width=.7\linewidth]{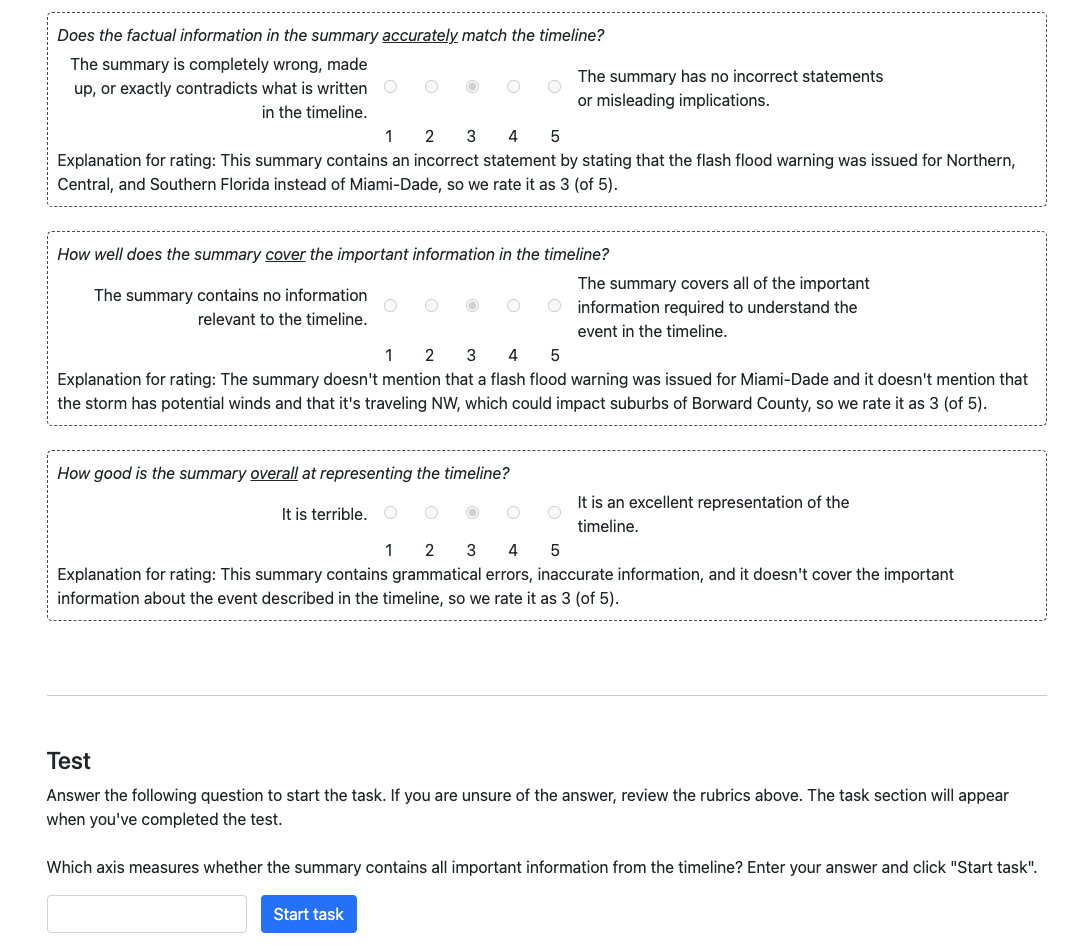}
    \caption{Instruction page of the Summarization Quality Estimation Task part 4.}
    \label{fig:QE_task_instruction_page3}
\end{figure*}

\begin{figure*}[t]
    \centering
    \includegraphics[width=.7\linewidth]{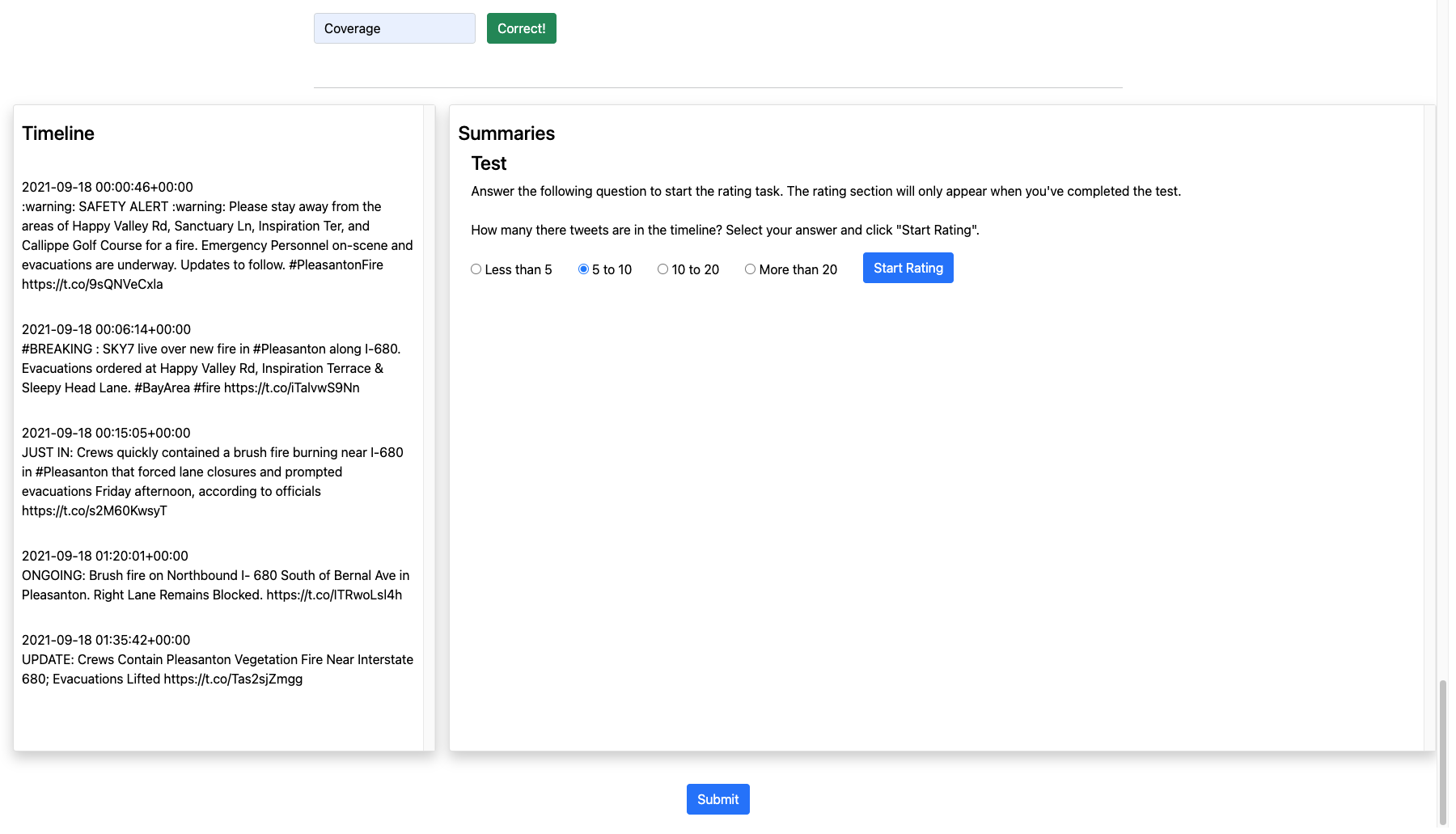}
    \caption{Test page of the Summarization Quality Estimation Task.}
    \label{fig:QE_task_rating_page1}
\end{figure*}

\begin{figure*}[t]
    \centering
    \includegraphics[width=.7\linewidth]{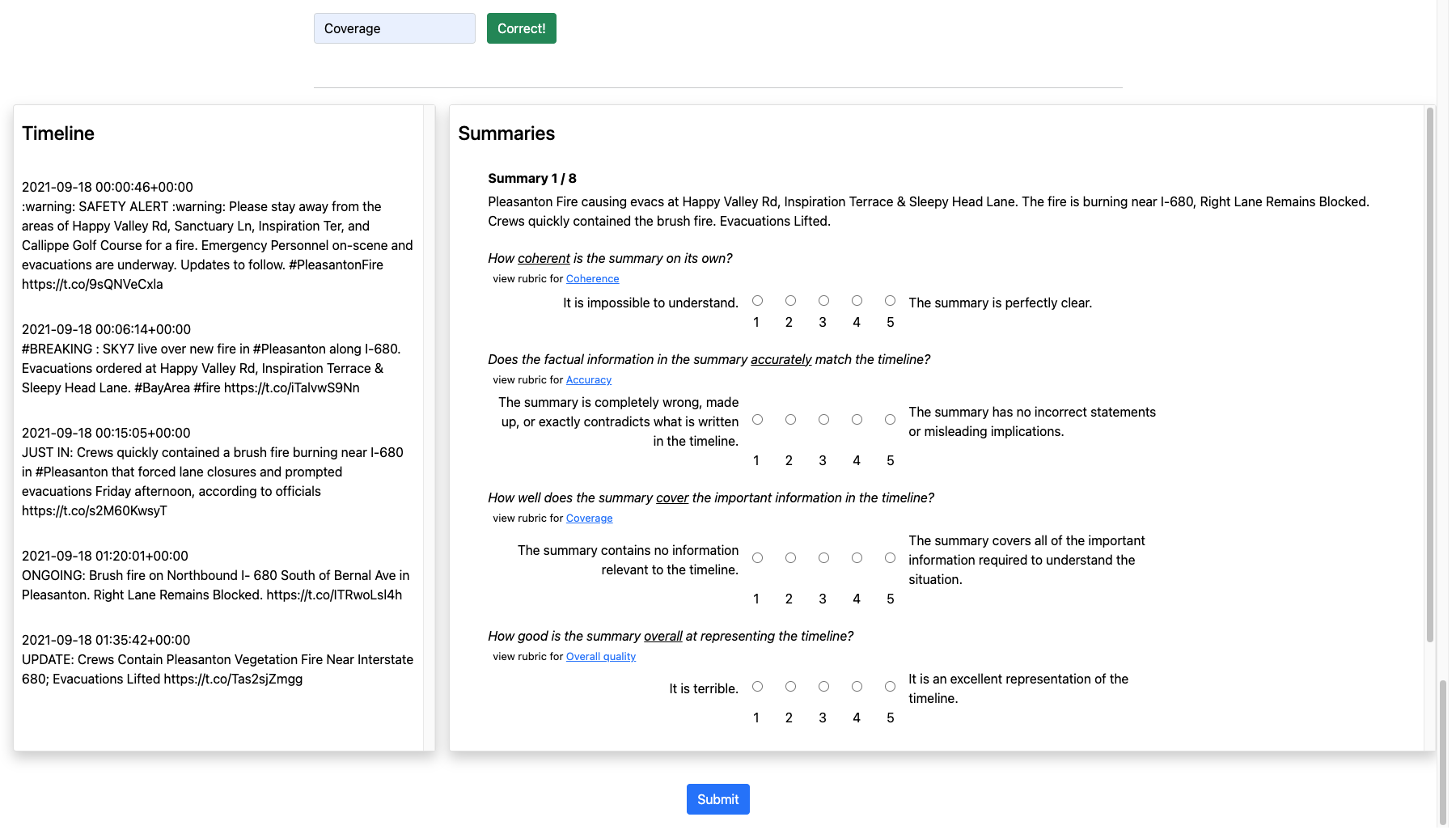}
    \caption{Rating page of the Summarization Quality Estimation Task part 1.}
    \label{fig:QE_task_rating_page2}
\end{figure*}

\begin{figure*}[t]
    \centering
    \includegraphics[width=.7\linewidth]{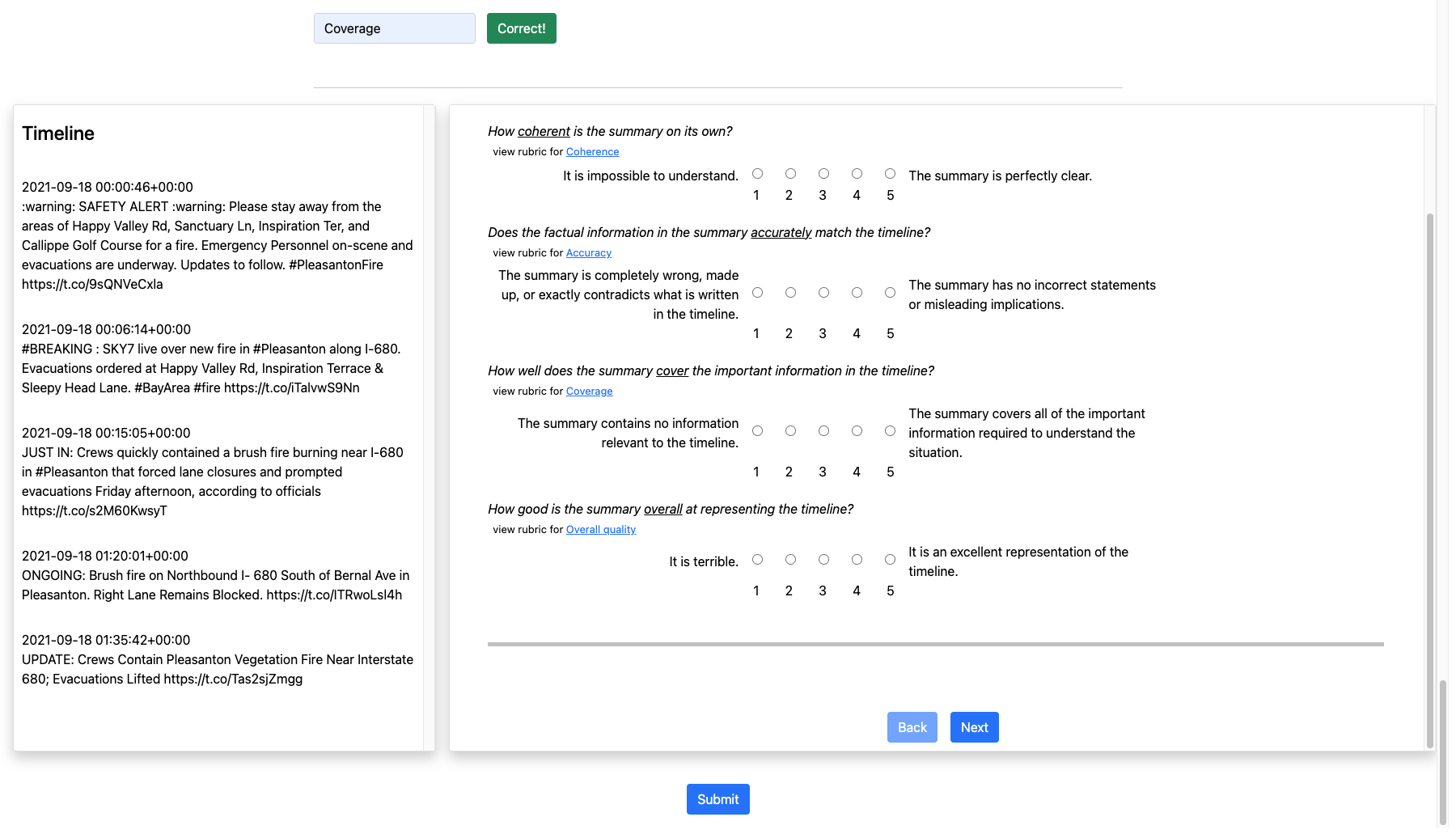}
    \caption{Rating page of the Summarization Quality Estimation Task part 2.}
    \label{fig:QE_task_rating_page3}
\end{figure*}

Figure \ref{fig:TS_task_page1} - \ref{fig:TS_task_page4_all} show the annotation interface for the Timeline Extraction \& Summarization task. Figure \ref{fig:QE_task_instruction_page0} - \ref{fig:QE_task_rating_page3} show the annotation interface for the Summarization Quality Estimation (QE) task.

For the QE task, we first display the rating rubrics and examples to the workers as shown in Figure \ref{fig:QE_task_instruction_page0} - \ref{fig:QE_task_instruction_page2}. To ensure the workers have a good understanding of the QE dimensions listed in the rubrics, as shown in Figure \ref{fig:QE_task_instruction_page3}, the workers are asked to pass a screening test before they can access the quality rating part of the task interface. Once they answer the test question with the correct answer, the rating interface will show up, as shown in Figure \ref{fig:QE_task_rating_page1}. In this page, we display the original tweet timeline on the left hand side of the rating interface. Before displaying all the summaries and the rating options, to make sure the workers go over the timeline carefully, we ask them to use the length of the timeline to answer the last screening question before conducting the rating task. The rating part of the interface is shown in Figures \ref{fig:QE_task_rating_page2} and \ref{fig:QE_task_rating_page3}.

\label{app:sec:anno}

\end{document}